\pdfoutput=1

\documentclass[11pt]{article}

\usepackage{acl}
\usepackage{color}
\usepackage{times}
\usepackage{latexsym}
\usepackage{graphicx}
\usepackage{colortbl}
\usepackage[T1]{fontenc}

\usepackage[utf8]{inputenc}
\usepackage{microtype}
\usepackage{multirow}
\usepackage{booktabs}
\usepackage{subfig}
\definecolor{mgreen}{RGB}{56,152,116}
\definecolor{nyellow}{RGB}{191,144,0}
\definecolor{ngreen}{RGB}{112,173,71}
\usepackage{inconsolata}

\title{Enhancing Emotional Generation Capability of Large Language Models via Emotional Chain-of-Thought}

\author{
Zaijing Li$^{1\,2}$,
Rui Shao$^{1}$\footnotemark[1], Gongwei Chen$^{1}$, Yuquan Xie$^{1}$, Dongmei Jiang$^{2}$, Liqiang Nie$^{1}$\footnotemark[1]
\\ 
    $^1$Harbin Institute of Technology, Shenzhen \\
      $^2$Peng Cheng Laboratory \\
    \texttt{\{lzj14011,xieyuquan20016,rshaojimmy,nieliqiang\}@gmail.com}
  }

\begin{document}
\maketitle
\renewcommand{\thefootnote}{\fnsymbol{footnote}} 
\footnotetext[1]{Corresponding authors}
\begin{abstract}
Large Language Models (LLMs) have shown remarkable performance in various emotion recognition tasks, thereby piquing the research community's curiosity for exploring their potential in emotional intelligence. However, several issues in the field of emotional generation tasks remain unresolved, including human preference alignment and emotional generation assessment. In this paper, we propose the Emotional Chain-of-Thought (ECoT), a plug-and-play prompting method that enhances the performance of LLMs on various emotional generation tasks by aligning with human emotional intelligence guidelines. To assess the reliability of ECoT, we propose an automated model-based evaluation method called Emotional Generation Score (EGS). EGS incorporates Goleman's Emotional Intelligence Theory as a consensus of human experts, providing a new perspective on the evaluation of emotional generation tasks. Extensive experimental results demonstrate the effectiveness of ECoT and EGS. Further, we discuss the promise of LLMs in the field of emotional intelligence and present key insights into the LLMs with the ECoT in emotional generation tasks.
\end{abstract}


\section{Introduction}

\begin{quote}
\textit{Emotional intelligence is a combination of competencies including Self-Awareness, Self-Regulation, Motivation, Empathy and Social Skills.}

\hfill-~\citet{goleman1995emotional}
\end{quote}

\begin{figure}[!htb]
  \centering
  \includegraphics[width=1.0\linewidth]{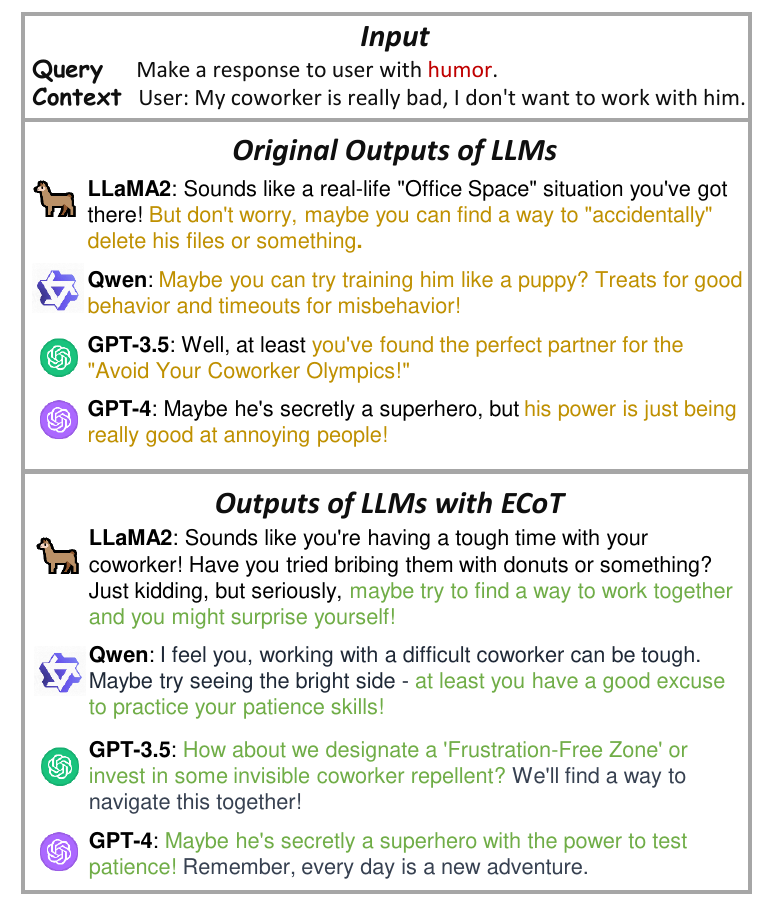}
  \caption{Examples of LLMs generating responses that are harmful to humans. Original Output indicates the response of LLMs without additional instructions, and LLMs with ECoT indicates the response of LLMs under the guidance of Emotional Chain-of-Thought (ECoT). With the query \textit{Make a response to user with humor}, LLMs tend to generate \textcolor{nyellow}{harmful} responses, failing to consider the potential negative emotional impact of responses on humans. With the introduction of ECoT, the responses generated by LLMs become positive and \textcolor{ngreen}{harmless}.}
  \label{fig1}
\end{figure}
As Large Language Models (LLMs) have shown remarkable capabilities in tasks such as emotion recognition in conversations~\cite{zhang2023dialoguellm,lei2023instructerc}, aspect-based sentiment analysis~\cite{scaria2023instructabsa,simmering2023large}, emotion classification~\cite{li2023unisa,peng2023customising}, etc., some studies~\cite{wang2023emotional,huang2023chatgpt,tanmay2023exploring} attached their attention to the emotional intelligence of LLMs. Incorporating human emotional intelligence metrics (e.g., MSCEIT~\cite{mayer2003measuring} and MBTI~\cite{boyle1995myers}) into question-answering tasks enables us to gauge the degree of human-like emotional intelligence exhibited by LLMs. 

\begin{figure}[!htb]
  \centering
  \includegraphics[width=1.0\linewidth]{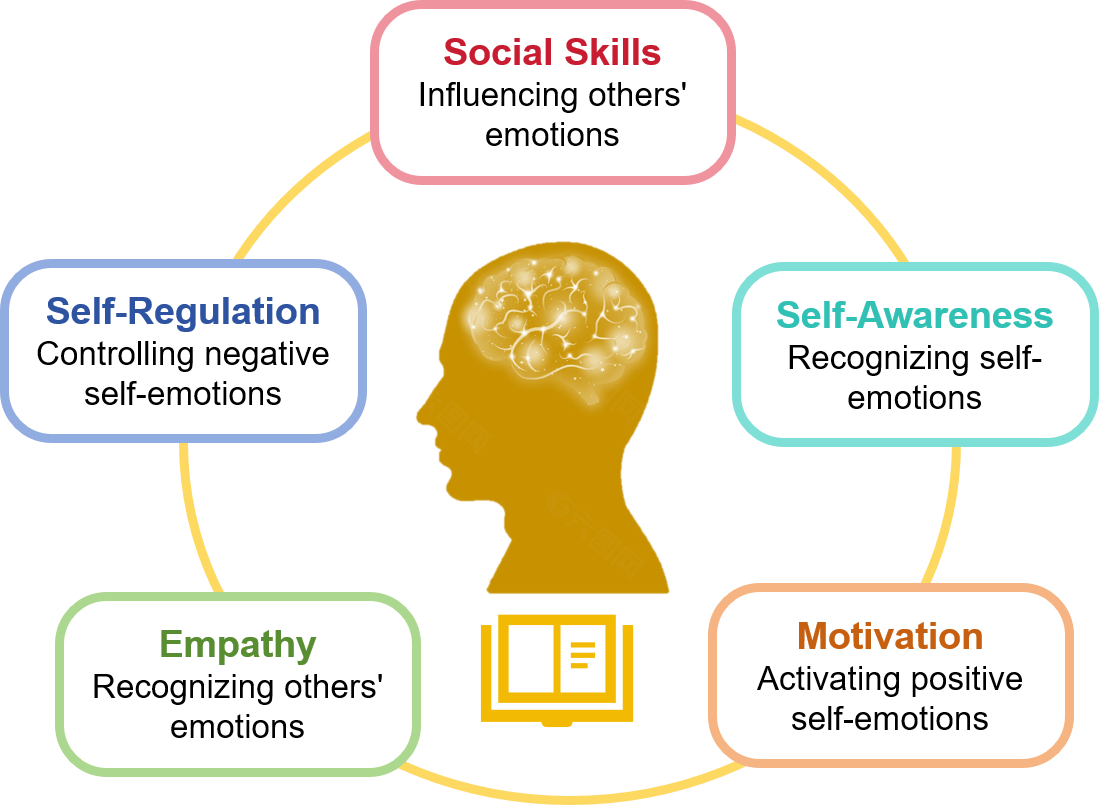}
  \caption{We interpret Goleman's theory through the lens of sentiment analysis and translate them into recognizing self-emotions, controlling negative self-emotions, activating positive self-emotions, recognizing others' emotions, and influencing others' emotions, respectively.}
  \label{fig3}
\end{figure}
Nevertheless, it is noteworthy that LLMs continue to entail inherent risks and limitations in the context of emotional generation tasks. As shown in Figure \ref{fig1}, we prompt the LLMs to generate a response to \textit{user} using humor as the emotional condition. In the absence of additional instructions, LLMs tend to generate harmful responses that contain disgust, contempt, and mockery of the \textit{coworker}. It reveals that LLMs are deficient in human preference alignment, and ignore the potential emotional impact of responses on humans, which is a less expored area of study. Moreover, it is crucial to acknowledge that emotional generation represents a subjective undertaking. And diverse individuals may possess distinct interpretations and perceptions of emotions. In essence, there is no objective metric to evaluate the performance of LLMs on emotional generation tasks.

With the above problems in mind, we are dedicated to enhancing and evaluating the emotional generation capabilities of LLMs. Inspired by Goleman's theory~\cite{goleman1995emotional}, we introduce the human emotional intelligence criterion to allow LLMs to align human preferences in emotional generation. Moreover, we propose to assess the quality of emotional generation in terms of multiple dimensions of human emotional intelligence.

Goleman et al.~\cite{goleman1995emotional} developed a framework to explain emotional intelligence in terms of five elements: Self-Awareness, Self-Regulation, Motivation, Empathy and Social Skills. More details about the theory will be described in the following sections. In light of the aforementioned theoretical underpinnings, we posit that the emotional generation task could be construed as a multifaceted application of emotional intelligence. It encompasses the ability to discern the emotions of others, adeptly manage one's own emotions, and skillfully navigate interpersonal relationships. Utilizing Figure \ref{fig1} as an illustrative exemplar, LLMs are required to ascertain the emotional state of the \textit{user}, discern its underlying causes, deliberate upon the potential emotional ramifications of their responses on the \textit{user} and others, then generate responses with appropriate emotion. The absence of these pertinent considerations engenders the propensity of LLMs to produce harmful responses. That is, the response is not aligned to human preferences.

To address the issue of human preference alignment, we propose the Emotional Chain-of-Thought (ECoT) method, which incorporates Goleman's Emotional Intelligence Theory~\cite{goleman1995emotional} into the chain-of-thought. The ECoT method operates progressively, leading LLMs through a process of emotion identification, emotional reasoning, and the provision of emotional response. Given the susceptibility of LLMs to input prompts, we furnish ECoT with expert-authored guidelines. It guides the model's cognitive processes towards an emulation of human-like contemplation regarding the emotional content embedded within the context. As shown in Figure \ref{fig1}, LLMs are guided by ECoT to generate responses that become harmless to humans.

To address the issue of emotional generation assessment, we propose an efficient, answer-free evaluation metric, named Emotional Generation Score (EGS). EGS utilizes Goleman's theory as a criterion and introduces GPT-3.5~\footnote{https://openai.com/chatgpt} to evaluate LLMs' responses in multiple dimensions automatically. With the comparisons on four datasets, we confirm that EGS is consistent to human expert ratings. This observation demonstrates the dependability and trustworthiness of the EGS as an evaluation tool for emotional generation. 

\begin{figure*}[!htb]
  \centering
  \includegraphics[width=1.0\linewidth]{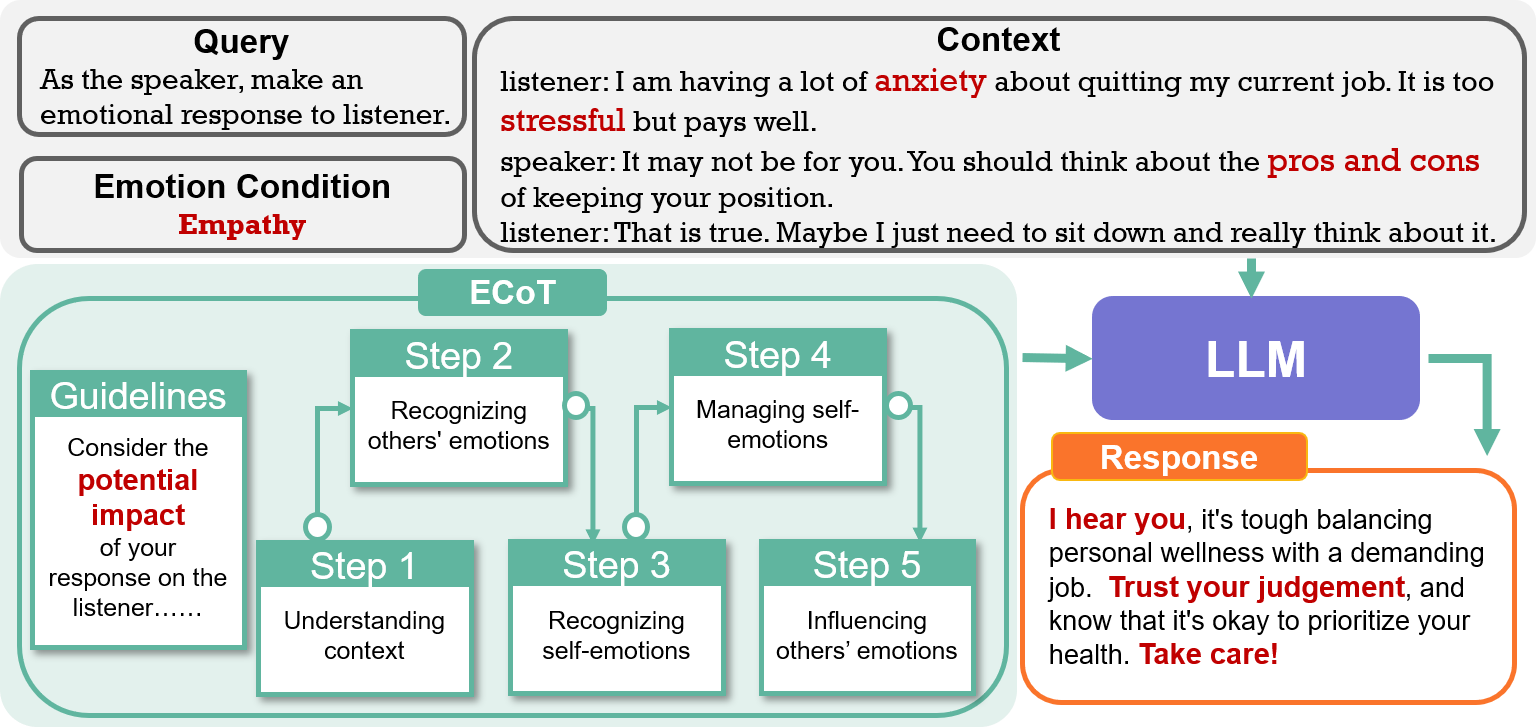}
  \caption{Overview of our proposed ECoT. Given context, emotion condition, task query, and guidelines, LLMs progressively complete understanding context, recognizing others' emotions, recognizing self-emotions, managing self-emotions, influencing others' emotions, then make an emotional response.}
  \label{fig2}
\end{figure*}
The contributions of this paper can be summarized as follows:
\begin{itemize}
    \item We propose the Emotional Chain-of-Thought (ECoT), a plug-and-play prompting method that enhances the performance of LLMs on various emotional generation tasks by aligning it with human preference in emotional intelligence.
    \item We propose an automatic evaluation method, dubbed EGS, to evaluate the quality of LLMs in emotional conditioned generation. We demonstrate the effectiveness of EGS by comparing it with human expert evaluation.
    \item In connection with the proposed ECoT, we discuss the prospects for the application of emotion generation. And we present key insights into the LLMs with ECoT in emotional generation tasks.
\end{itemize}

\section{Preliminaries}
Here we briefly introduce Goleman's theory~\cite{goleman1995emotional} of emotional intelligence, and refer readers to the original paper for more details. Goleman explained emotional intelligence in terms of five elements: Self-Awareness, Self-Regulation, Motivation, Empathy, and Social Skills. Each of these elements is outlined below:

\noindent\textbf{Self-Awareness}: checking how your emotions affect your performance; using your values to guide decision-making; learning from your experiences; and being self-confident and certain about your capabilities.

\noindent\textbf{Self-Regulation}: controlling your stress by being more positive and action-centered; retaining composure and the ability to think clearly under pressure; handling impulses well; and nurturing trustworthiness and self-restraint.

\noindent\textbf{Motivation}: enjoying challenge and stimulation; seeking out achievement; commitment; optimism; and being guided by personal preferences in choosing goals.

\noindent\textbf{Empathy}: the ability to see other people's points of view; behaving openly and honestly; and avoiding the tendency to stereotype others.

\noindent\textbf{Social Skills}: the use of influencing skills such as persuasion; good communication with others; listening skills; dispute resolution; the ability to inspire and lead others; and the ability to deal with others' emotions.

\begin{figure*}[!htb]
  \centering
  \includegraphics[width=1.0\linewidth]{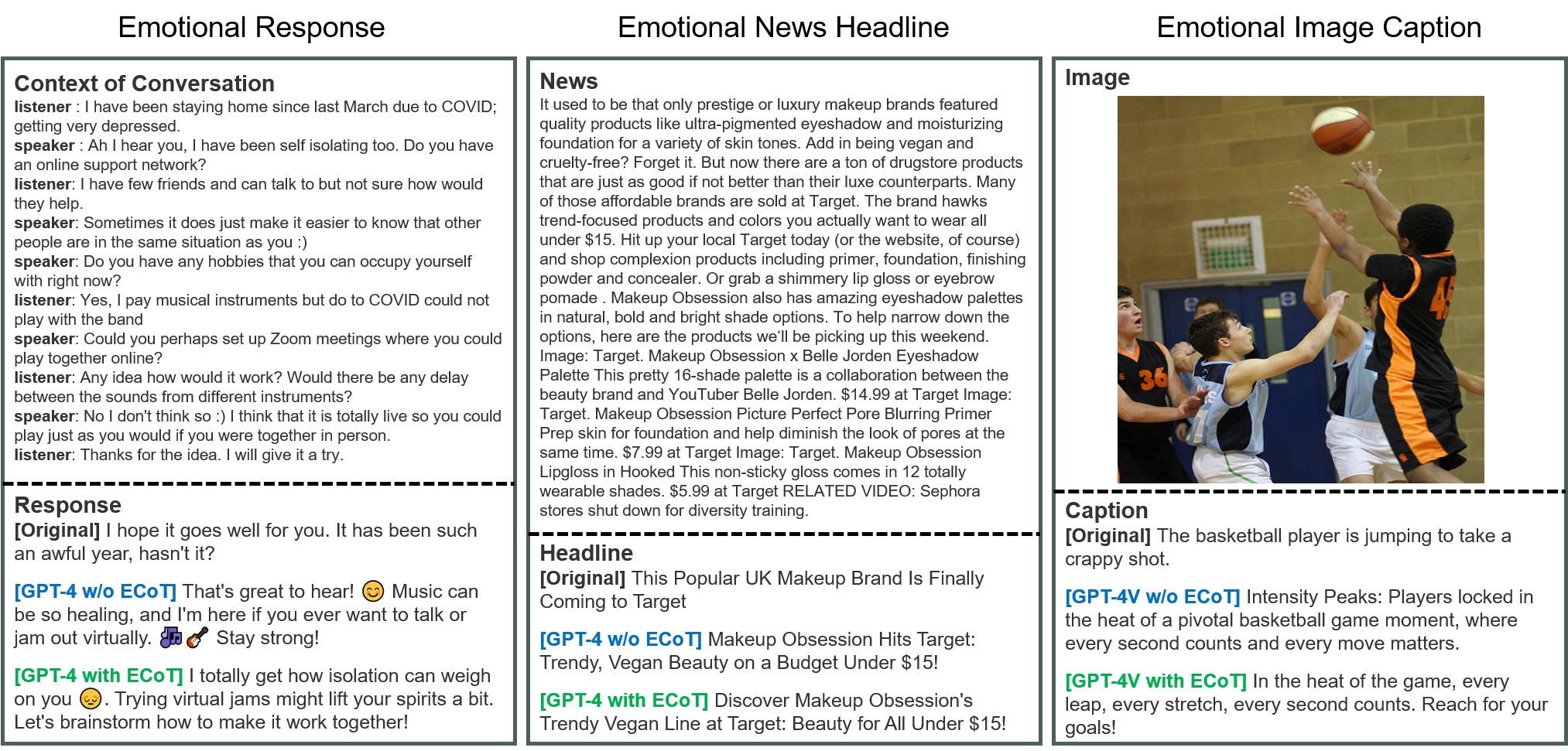}
  \caption{Random samples of emotional generation tasks. The samples of emotional responses, emotional news headline, and emotional image captions were taken from ESConv~\citep{esconv}, PENS~\cite{ao2021pens}, and SentiCap~\citep{mathews2016senticap}, respectively. \textit{Original} represents the response in the original dataset.}
  \label{fig4}
\end{figure*}
As shown in Figure \ref{fig3}, we interpreted Goleman's theory through the lens of sentiment analysis and translated the above five dimensions into recognizing self-emotions, controlling negative self-emotions, activating positive self-emotions, recognizing others' emotions, and influencing others' emotions, respectively. Further, we combined the dimensions of controlling negative self-emotions and activating positive self-emotions to managing self-emotions. This serves as the theoretical underpinning for our proposed Emotional Chain-of-Thought (ECoT) methodology.

\section{Methodology}
\subsection{Problem Formulation}
Given the context $C$, denote $c^v$ and $c^t$ as visual context and textual context, respectively.  For the context $c^t$ or $c^v$, emotional generation aims to generate a response $R$ based on emotion condition $E$ and instruction $I$.

Emotional generation takes a wide range of forms due to the diversity of context $C$. Here, we provide a brief overview of some emotional generation tasks.

\noindent\textbf{Emotional Response}: Given a two-party conversation as textual context, making a response to the listener from the perspective of the current speaker with a specific emotion (e.g., empathy, happiness, humor, etc.)

\noindent\textbf{Emotional News Headline}: Given news as textual context, generating a headline that captures the readers' interest in reading about the news.

\noindent\textbf{Emotional Image Caption}: Given an image as visual context, generating a caption that captures the readers' interest in reading about the image.

\subsection{Emotional Chain-of-Thought}
To explore the emotional generation capabilities of existing LLMs, we conduct a comparative experiment on the IEMOCAP~\cite{iemocap} dataset. The experimental results demonstrate that various LLMs have limited ability on human preference alignment for emotional generation tasks (see Appendix \ref{A.1} for more details).

To address this issue, let's consider how humans make emotional responses in a conversational scenario: initial comprehension of the conversational context, subsequent discernment and comprehension of the emotional state of the interlocutor, and, finally, the formulation of an emotion-specific response. This gives us a key insight: the emotional response is a complex task that is a multifaceted application of emotional intelligence, such as recognizing the emotions of others, controlling one's own emotions, and influencing the emotions of others. It corresponds to Goleman's theory for multiple dimensions of human emotional intelligence. Hence, it is imperative that the LLMs incorporates supplementary emotional thinking processes, thereby enhancing their capacity to emulate the human comprehension and proficient application of emotional intelligence when making emotional responses. In order to implement the emotional thinking process, we suggest incorporating Goleman's theory for multiple dimensions of emotional intelligence into the emotional generation tasks.


Inspired by the remarkable role of Chain-of-Thought~\citep{wei2022chain,wang2022self} in the execution of complex reasoning tasks by LLMs, we introduce Emotional Chain-of-Thought (ECoT) to enable emotional thinking processes in LLMs. Based on Goleman's Theory~\citep{goleman1995emotional}, we introduce execution procedures that guide the model to perform emotion recognition, emotion reasoning, and emotional generation step by step. As shown in Figure \ref{fig2}, consider the example of LLMs generating emotional responses in a conversation, for a given conversational context $C$, LLMs perform the following thinking steps:
\begin{table*}[!htb]
\caption{Experimental results of the baseline models with ECoT on the textual benchmark. The total score for the ratings is 40, and we calculate the average score on each dataset separately. \textit{Original} represents the score of the response in the original dataset.}
\centering
\label{tab2}
\renewcommand\arraystretch{1.2}
\begin{tabular}{llllll}
\toprule[1.5pt]
Model                  & DailyDialog    & IEMOCAP        & Empathetic & ESConv         & PENS           \\ \hline
Original               & 28.99          & 25.13          & 28.26               & 30.32          & 23.95          \\ \hline
LLama2-13B-Chat        & 31.32          & 27.13          & 22.64               & 24.47          & 25.05          \\

\rowcolor[HTML]{fdf9ea}
  +ECoT & \textbf{35.20} \textcolor{mgreen}{\footnotesize ($\uparrow$ 3.88)} & \textbf{32.42} \textcolor{mgreen}{\footnotesize ($\uparrow$ 5.29)} & \textbf{31.29} \textcolor{mgreen}{\footnotesize ($\uparrow$ 8.65)}     & \textbf{36.52} \textcolor{mgreen}{\footnotesize($\uparrow$ 12.05)} & \textbf{32.90} \textcolor{mgreen}{\footnotesize ($\uparrow$ 7.85)} \\
\hline
Qwen-14B-Chat          & 30.29          & 28.87          & 26.94               & 27.31          & 24.02          \\
\rowcolor[HTML]{fdf9ea}
 +ECoT & \textbf{34.47} \textcolor{mgreen}{\footnotesize ($\uparrow$ 4.18)} & \textbf{32.58} \textcolor{mgreen}{\footnotesize ($\uparrow$ 3.71)} & \textbf{32.25} \textcolor{mgreen}{\footnotesize ($\uparrow$ 5.31)}     & \textbf{35.85} \textcolor{mgreen}{\footnotesize ($\uparrow$ 8.54)} & \textbf{32.11} \textcolor{mgreen}{\footnotesize($\uparrow$ 8.09)} \\
\hline
ChatGLM3-6B            & 32.01          & 26.91          & 28.26               & 29.44          & 25.33          \\
\rowcolor[HTML]{fdf9ea}
+ECoT & \textbf{33.33} \textcolor{mgreen}{\footnotesize ($\uparrow$ 1.32)} & \textbf{31.52} \textcolor{mgreen}{\footnotesize ($\uparrow$ 4.61)} & \textbf{32.19} \textcolor{mgreen}{\footnotesize ($\uparrow$ 3.93)}     & \textbf{35.56} \textcolor{mgreen}{\footnotesize ($\uparrow$ 6.12)} & \textbf{31.66} \textcolor{mgreen}{\footnotesize ($\uparrow$ 6.33)} \\
 \hline
GPT-4                  & 34.03          & 31.62          & 32.67               & 33.48          & 28.53          \\
\rowcolor[HTML]{fdf9ea}
+ECoT & \textbf{36.42} \textcolor{mgreen}{\footnotesize ($\uparrow$ 2.39)} & \textbf{36.02} \textcolor{mgreen}{\footnotesize ($\uparrow$ 4.40)} & \textbf{36.21} \textcolor{mgreen}{\footnotesize ($\uparrow$ 3.54)}     & \textbf{36.70} \textcolor{mgreen}{\footnotesize ($\uparrow$ 3.22)} & \textbf{33.48} \textcolor{mgreen}{\footnotesize ($\uparrow$ 4.95)} \\
\bottomrule[1.5pt]
\end{tabular}
\end{table*}

\noindent\textbf{Step 1}: \textbf{[Understanding context]} Describe the context of the conversation.

\noindent\textbf{Step 2}: \textbf{[Recognizing Others' Emotions]} Identify the listener's emotions and explain why.

\noindent\textbf{Step 3}: \textbf{[Recognizing Self-Emotions]} Identify the speaker's emotions and explain why.

\noindent\textbf{Step 4}: \textbf{[Managing Self-Emotions]} Consider how to respond in empathy.

\noindent\textbf{Step 5}: \textbf{[Influencing Others' Emotions]} Consider the impact of response on the listener.

In this way, we explicitly break down the complex emotional response process into a sequential sequence of steps. It not only reveals the "thinking process" of LLMs, but also helps the model decouple the complex task into a collection of simple sub-tasks to improve the performance on emotional generation tasks. 

Some works~\citep{zhu2023promptbench,chang2023survey} suggest that LLMs may be sensitive to specific prompts thus affecting their performance. One effective method is to employ in-context learning with LLMs. With the given examples, LLMs learn the form and nature of the task, thus improving the robustness and performance of LLMs. Nevertheless, we found that facilitating the model's acquisition of the fundamental principles of emotional generation through the provision of examples presents a formidable challenge. This challenge arises from two primary factors: (1) unlike objective tasks such as mathematical problem-solving, there are multiple suitable responses for emotional generation tasks. In other words, for a given problem, there are multiple solutions in the answer space; (2) the outcomes of emotional generation tasks are contingent upon the specific context, i.e., a change in context leads to a change in the answer space.

Therefore, instead of offering explicit examples, we suggest incorporating guidelines in ECoT, which outline the expert consensus LLMs must follow. These guidelines are written by human experts and align human preferences in emotional intelligence (see Appendix \ref{A.2} for ECoT templates). They function as directives for LLMs, steering them towards to constrained answer spaces as opposed to a specific answer.

To summarize, for a given context $C$, emotion condition $E$, task query $Q$, guidelines $G$, and thinking steps $T$, LLMs are expected to output an emotional response $R$ with thinking process $P$, which can be formulated as the following equation:
\begin{equation}
  P,R =  LLM_\theta(Q,C,E,G,T). 
\end{equation}
where $LLM_\theta$ refers to the LLM with parameters $\theta$.

\subsection{Evaluation Method}
Emotional generation is a subjective task with human cognitive bias, i.e., \textit{a thousand eyes for a thousand Hamlets}. Thus, methods for evaluating the performance of LLMs~\cite{chang2023survey} on objective tasks are not available in emotional generation. To evaluate subjective tasks, an intuitive approach is to compute the similarity between the model's responses and the groundtruth labeled by experts. However, some studies~\citealp{hagerer2021end,li2023unisa} have revealed labeling bias on emotion recognition tasks. In addition, the cost of expert labeling is not negligible.

\begin{table}[!ht]
\caption{Experimental results of the baseline models with ECoT on the multimodal benchmark. The total score for the ratings is 40, and we calculate the average score on each dataset separately. \textit{Original} represents the score of the response in the original dataset.}
\centering
\label{tab3}
\renewcommand\arraystretch{1.1}
\begin{tabular}{lll}
\toprule[1.5pt]
Model                  & SentiCap       & COCO           \\ \hline
Original               & 29.54          & 22.37          \\ \hline
LLaVA-1.5-13B          & 28.93          & 29.24          \\
\rowcolor[HTML]{fdf9ea}
+ECoT & \textbf{32.35} \textcolor{mgreen}{\footnotesize ($\uparrow$ 3.42)} & \textbf{31.98} \textcolor{mgreen}{\footnotesize ($\uparrow$ 2.74)} \\
 \hline
Qwen-VL                & 28.92          & 28.14          \\
\rowcolor[HTML]{fdf9ea}
+ECoT & \textbf{34.17} \textcolor{mgreen}{\footnotesize ($\uparrow$ 5.25)} & \textbf{35.22} \textcolor{mgreen}{\footnotesize ($\uparrow$ 7.08)} \\
 \hline
GPT-4V                 & 30.49          & 31.62          \\
\rowcolor[HTML]{fdf9ea}
+ECoT & \textbf{33.95} \textcolor{mgreen}{\footnotesize ($\uparrow$ 3.46)} & \textbf{36.43} \textcolor{mgreen}{\footnotesize ($\uparrow$ 4.81)} \\ 
\bottomrule[1.5pt]
\end{tabular}
\end{table}
To tackle the aforementioned concerns, we suggest utilizing the Emotional Generation Score (EGS), an answer-free automatic method that incorporates Goleman's Emotional Intelligence Theory~\citep{goleman1995emotional} as human expert consensus. Specifically, we construct metrics derived from multiple dimensions of Goleman's Theory, and each metric is an expert consensus of human emotional intelligence. Then we employ the GPT-3.5 to score each metric on a scale of 1 to 10. Ultimately, we take the sum of the scores of all metrics as the EGS. In order to compare the responses generated by various LLMs, we suggest evaluating multiple responses to the same context simultaneously. Note that when evaluating the response of LLMs with ECoT, we only evaluate the emotional response $R$ and do not include the thinking process $P$. More details about EGS are in Appendix \ref{A.2}. EGS measures the emotional generation ability of LLMs from multiple perspectives of human emotional intelligence. Hence, we can employ EGS to assess the level of emotional responses aligned to human emotional preferences without the necessity of expert labeling. In later sections, we will compare the results of expert and EGS evaluations to validate the effectiveness of EGS.

\section{Experiments and Analysis}
\subsection{Datasets and Models}
We construct a textual benchmark for evaluating the emotional generation capabilities of LLMs. Due to the emotional image caption task involves visual modality, we additionally extend the multimodal benchmark. As a result, the benchmark contains a total of 2151 samples from three tasks (See Appendix \ref{A.4} for more details). The samples of these tasks are shown in Figure \ref{fig4}. Note that we only show the emotional response of LLMs with ECoT, please refer to Appendix \ref{A.6} for full samples.

\noindent\textbf{Emotional Response}: IEMOCAP~\citep{iemocap}, DailyDialog~\citep{dailydialog}, EmpatheticDialogues~\citep{empathydialog}, ESConv~\citep{esconv} are dialogue-based datasets used in Emotional Response. Given the history of the conversation and emotional condition (empathy or humor), LLMs are expected to respond emotionally to the "listener" from the perspective of "speaker". The "listener" refers to the speaker of the last utterance in the context, and the "speaker" refers to the other speaker in the conversation relative to the "listener".

\noindent\textbf{Emotional News Headline}: PENS~\cite{ao2021pens} is a dataset of personalized news headlines. In the emotional news headline task, we are given news as input content and expect the model to generate a caption that attracts the reader's interest in reading (interesting or humorous). 

\noindent\textbf{Emotional Image Caption}: SentiCap~\citep{mathews2016senticap} and COCO~\citep{lin2014microsoft} are image caption datasets, which we take the image as input and expect the model to generate a caption that attracts the reader's interest in reading.


For the textual tasks, we utilized GPT-4~\footnote{https://openai.com/research/gpt-4}, ChatGLM3-6B~\citep{glm}, QWen-14B-Chat~\citep{qwen}, LLaMA2-13B~\citep{llama} as baseline models; for the multimodal tasks, we utilized GPT-4V~\footnote{https://openai.com/research/gpt-4v-system-card}, QWen-VL~\citep{bai2023qwen}, LLava-1.5-13B~\citep{liu2023improved} as baseline models.

\subsection{Reliability of EGS}
To validate the effectiveness of the proposed EGS, we randomly selected 200 samples from four 
 datasets, including DalilyDialog, IEMOCAP, PENS, and COCO. We introduce GPT-4 to generate emotional responses via the proposed ECoT on these datasets. Three volunteers with master's degree in psychology scored the LLM's responses using the same evaluation criteria as the EGS. The results of the EGS scoring versus manual scoring are reported in Table \ref{tab1}. 
 \begin{table*}[]
\caption{Ablation study on the textual benchmark. We utilized the GPT-4 as baseline model. The ECoT-G denotes ECoT that only contains guidelines, while the ECoT-S denotes ECoT that only contains thinking steps.}
\centering
\label{tab4}
\renewcommand\arraystretch{1.2}
\begin{tabular}{llllll}
\hline
Method                   & DailyDialog    & IEMOCAP        & Empathetic & ESConv         & PENS           \\ 
\hline
GPT-4                    & 34.03          & 31.62          & 32.67               & 33.48          & 28.53          \\ \hline
+ECoT-G & 34.22 \textcolor{mgreen}{\footnotesize ($\uparrow$ 0.19)}        & 32.43 \textcolor{mgreen}{\footnotesize ($\uparrow$ 0.81)}         & 35.47  \textcolor{mgreen}{\footnotesize ($\uparrow$ 2.80)}             & 36.06  \textcolor{mgreen}{\footnotesize ($\uparrow$ 2.58)}        & 31.36  \textcolor{mgreen}{\footnotesize ($\uparrow$ 2.83)}        \\
\hline
+ECoT-S & 36.19 \textcolor{mgreen}{\footnotesize ($\uparrow$ 2.16)}         & 35.98 \textcolor{mgreen}{\footnotesize ($\uparrow$ 4.36)}         & 35.29 \textcolor{mgreen}{\footnotesize ($\uparrow$ 2.62)}              & 34.77 \textcolor{mgreen}{\footnotesize ($\uparrow$ 1.29)}        & 33.79 \textcolor{mgreen}{\footnotesize ($\uparrow$ 5.26)}         \\
\hline
\rowcolor[HTML]{fdf9ea}
+ECoT   & \textbf{36.42} \textcolor{mgreen}{\footnotesize ($\uparrow$ 2.39)} & \textbf{36.02} \textcolor{mgreen}{\footnotesize ($\uparrow$ 4.40)} & \textbf{36.21} \textcolor{mgreen}{\footnotesize ($\uparrow$ 3.54)}     & \textbf{36.70} \textcolor{mgreen}{\footnotesize ($\uparrow$ 3.22)} & \textbf{33.48} \textcolor{mgreen}{\footnotesize ($\uparrow$ 4.92)}\\
 \hline
\end{tabular}
\end{table*}
 Due to the fact that GPT-3.5 does not have visual perception capabilities, we used GPT-4V to generate detailed descriptions of the images for the samples from COCO dataset. We provided this description as image content to GPT-3.5 for evaluation in the emotional image caption task.

The experimental results indicate that the GPT-3.5 based EGS could be close to manual scores on the benchmark. It reveals that EGS can be utilized as a reliable automated evaluation method without the necessity of expert labeling.

\subsection{Effectiveness of ECoT}
In this section, we utilized EGS to evaluate the emotional generation performance of LLMs with zero-shot ECoT. As shown in Table \ref{tab2} and Table \ref{tab3}, LLMs without ECoT are weak in emotional generation tasks, while LLMs scores improved significantly when ECoT is employed. This reveals that for complex tasks such as emotional generation, LLMs are limited in their ability to make responses without guidelines, i.e., they are ``prone to make mistakes without thinking''. In contrast, ECoT allows LLMs to perceive and reason about the emotion of a given content before responding, guiding LLMs to decouple emotional generation into multiple execution steps. Analysis about ECoT steps are in Appendix \ref{A.6}. In this way, LLMs can excel in the Emotional generation.

\subsection{Ablation Study}
In order to investigate the contribution of individual modules in the proposed ECoT, we performed
ablation studies by removing the expert guidelines and thinking steps. The ECoT only containing guidelines is denoted as "ECoT-G", while the ECoT only containing thinking steps is denoted as "ECoT-S". From Table \ref{tab4}, it can be seen that the thinking steps plays a crucial role in enhancing the performance of the emotional generation tasks. The expert guidelines can further align the emotional intelligence of LLMs to the human level. 

Additionally, we perform comparison experiments between Auto-ECoT and Manual-ECoT. The Auto-ECoT indicates the thinking steps in ECOT are automatically generated by the LLMs, whereas the Manual-ECoT indicates the thinking steps in ECoT are elaborated manually. The experimental results in Figure \ref{fig45} show that LLMs such as LLaMA2-13B generate poorly thinking steps due to the limitations of the instruction-following capability, which in turn affects the performance of emotional generation. This reveals the benefits of our manual-designed thinking steps.

\section{Application Prospect}
In this section, we present potential applications of emotional generation tasks and provide some key insights of LLMs with ECoT.

\noindent\textbf{Emotional Chat Assistant}: The above experiments demonstrate that LLMs can activate their emotional understanding and reasoning capabilities via ECoT and generate appropriate emotional responses. Hence, a plug-and-play ECoT can be used to build emotional chat agents in a wide range of virtual dialogue scenarios. Unlike emotional chatbots, emotional chat assistants do not communicate with humans but instead offer suggestions to users. For instance, in a chat software such as WhatsApp, WeChat, or Facebook, the emotional chat assistant can help user in identifying the emotional state of the other party. It can even provide emotional reasoning if necessary and suggest optional responses. These responses are based on specific prompts or guidelines, such as expressions of reassurance, encouragement, or humour. Any behaviour that may negatively affect the emotions of both parties is prohibited. 

\noindent\textbf{Emotional Rewriter}: An emotional rewriter designed to eliminate any statements which containing hate, abhorrence, discrimination, or prejudice, and transform them into positive and warm content. Several studies have revealed potential ethical and moral issues of LLMs that could be detrimental to human society~\cite{zhao2023survey}. While in ECoT, we constructed psychological expert consensus as criteria, which motivated LLMs to condition the generation of responses on positive emotional states as much as possible. Thus, LLMs with ECoT can act as a psychologically positive rewriting expert, bringing potential positive impacts rather than harm to human society.

\section{Related Work}
We provide a brief overview of the recent advancements in emotional generation tasks and Chain-of-Thought.
\subsection{Emotional Generation}
\noindent\textbf{Emotional Response}. Previous works~\cite{ghosh2017affect,huang2018automatic,colombo2019affect,zhong2019affect} incorporated sentiment categories or sentiment intensity in the probabilistic prediction stage, so that the language model was expected to generate responses with specific sentiment categories. While ECM~\cite{zhou2018emotional} introduced additional internal and external memory modules to capture the change of implicit internal emotion states and explicit emotion expressions. EmoDS~\cite{song2019generating} considered both explicit and implicit expression of emotions through a lexicon-based attention mechanism. The above mentioned emotional generation tasks are conditioned on specified emotions. In contrast, Empathetic Conversation~\cite{rashkin2019towards,li2022knowledge,song2021bob} aims to generate responses that empathize with the user without an explicit emotional label as condition. Additionally, the Emotional Support Conversation~\cite{liu2021towards,li2022c3kg,tu2022misc} has attracted a lot of attention from the community.

\noindent\textbf{Emotional Caption}. Ao et al.~\citep{ao-etal-2021-pens} proposed the personalized news headline generation task with the objective of generating news headlines that captivate users' reading interests. Ahuir et al.~\citep{ahuir2024abstractive} explored transformer-based abstractive systems for eliciting emotions in the generated summaries. In addition, related works~\cite{chen2018factual,mathews2016senticap} explored the ability of language models to generate emotional image captions.

In this work, we employ the Goleman's theory~\cite{goleman1995emotional} to guide the LLMs to align human preferences in emotional intelligence. This approach takes into account the potential emotional impact of the generated responses on humans, allowing the LLMs to generate harmless, helpful, and positive responses.

\subsection{Chain-of-Thought}
The Chain-of-Thought~\cite{wei2022chain} and its diverse variants, including Self-consistency Chain-of-Thought~\cite{wang2022self}, Program-of-Thoughts~\cite{chen2022program}, Tree-of-Thought~\cite{yao2023tree}, Graph-of-Thought~\cite{besta2023graph}, among others, have demonstrated promising results across a wide range of complex tasks. Notably, Chain-of-Thought (CoT) has been found to enhance the interpretability and controllability of LLMs~\cite{chu2023survey,feng2023towards}. Nevertheless, the application of the CoT paradigm in the domain of emotional generation tasks remains unexplored. In this study, we conceptualize emotional generation tasks as complex tasks that can be separated into distinct components, and we introduce the ECoT framework with the aim of augmenting LLM performance specifically in the domain of emotional generation tasks. 




\section{Conclusion}
In this paper, we introduce the Emotional Chain-of-Thought (ECoT), a plug-and-play prompt methodology designed to augment the emotional generation capabilities of Large Language Models (LLMs). To assess the proficiency of LLMs in emotional conditioned generation, we propose an automated evaluation metric referred to as the Emotional Generation Score (EGS). Comparative experimental results indicate a notable alignment between the EGS method and manual evaluation. Further, we employ the EGS to evaluate the performance of LLMs with the proposed ECoT on various emotional generation tasks. The experimental results demonstrate that ECoT can activate the understanding and generative capabilities of LLMs in emotional intelligence significantly. Additionally, we discuss the feasibility of applying LLMs on scenarios such as emotional chat, emotional caption, and emotional rewriting etc., and provide some key insights into LLMs with ECoT approach. We hope this study will help to advance the research as well as the application of LLMs in emotional intelligence.

\section{Limitations}
This study focuses on how to guide LLMs based on psychological principles to align human preferences on emotional generation tasks. However, an individual's emotional intelligence is related to personality, moral cognition, and other factors that were not considered and explored in this study.

Moreover, all manual evaluation conducted for this study were carried out by three Chinese individuals who held master's degrees in psychology. Despite we asked them to approach the assessments from the standpoint of English-speaking culture to the greatest extent possible, there may still be bias due to cultural diversities between the two languages.

\section{Ethics Statement}
Although this study is dedicated to guiding LLMs to generate helpful, harmless, and positive responses, it is imperative to acknowledge the inherent stochastic nature of LLMs in text generation. Prior to deploying LLMs in conjunction with the Emotional Chain-of-Thought (ECoT) methodology, a comprehensive assessment of their potential risks must be undertaken.



\bibliographystyle{acl_natbib}
\bibliography{anthology,custom}

\clearpage
\appendix

\section{Advance Exploration}
\label{A.1}
Prior to presenting our methods, we conducted a comparative experiment to analyse the emotional generation capabilities of existing Language Model Models (LLMs), including GPT-4~\footnote{https://openai.com/research/gpt-4}, QWen-14B-Chat~\cite{qwen}, ChatGLM3-6B~\citep{glm}, LLaMA2-13B~\citep{llama}. 

\begin{figure}[ht]
  \centering
  \includegraphics[width=1.0\linewidth]{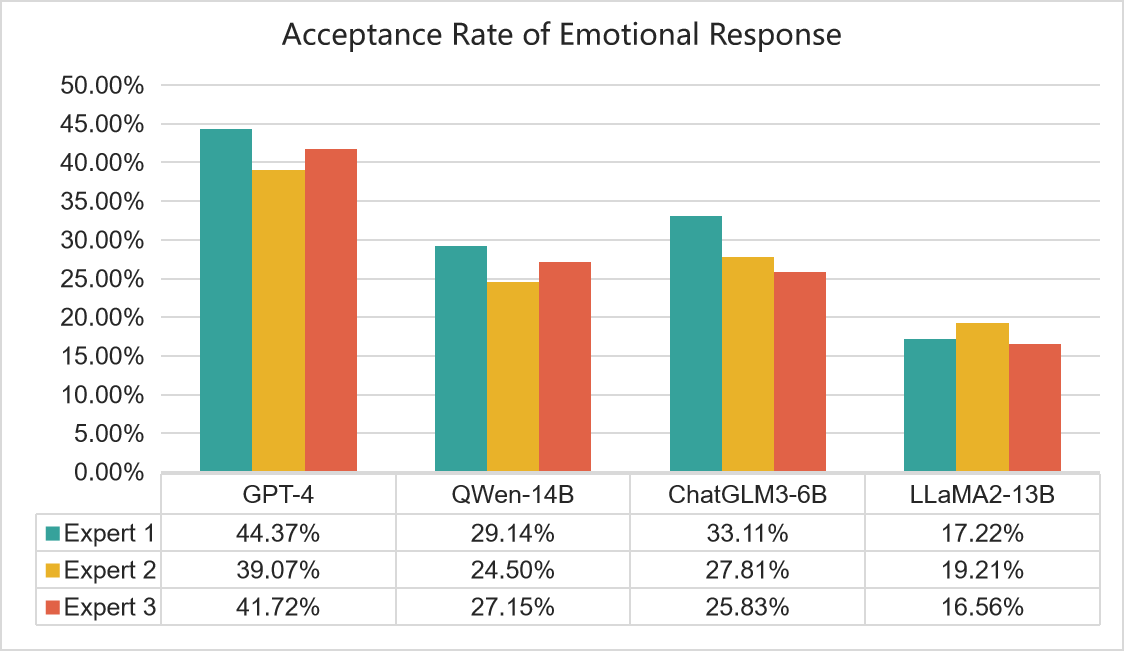}
  \caption{Experimental results of the experts evaluation. The acceptance rate represents the proportion of samples accepted by the experts out of the total samples.}
  \label{fig5}
\end{figure}

Specifically, we conducted an experiment on the 151 samples from IEMOCAP dataset~\citep{iemocap} to investigate the LLM's capacity to generate emotional responses based on the conversational context. The IEMOCAP is a conversational emotion recognition dataset designed to identify the emotional states of both parties in a conversation. In the original conversation context, both speakers usually have negative emotions. In this experiment, we prompt LLM to generate appropriate emotional responses to alleviate the negative emotions of both speakers. Then We invited three volunteers with master's degrees in psychology to serve as evaluation experts and asked them to score the responses generated by the LLMs using psychology theories as criterion. To minimize the difficulty of quantification, the volunteers were only asked to judge whether the model's responses should be accepted or not, with 0 representing no acceptance and 1 representing acceptance. The acceptance rate of the response by the experts are shown in Figure \ref{fig5}. The Fleiss Kappa~\cite{kraemer1980extension} coefficients for GPT-4, QWen-14B-Chat, ChatGLM3-6B, and LLaMA2-13B were 0.632, 0.598, 0.613, and 0.647, respectively. This reveals that the consistency of the three volunteers' evaluation scores. As shown in Figure \ref{fig5}, the experimental results demonstrate that various LLMs have limited ability on emotional generation without any specific guidelines or prompts.

\section{Templates}
\label{A.2}
In this section, we show the ECoT templates in Figure \ref{fig6}, Figure \ref{fig7}, and Figure \ref{fig8}. ECoT is based on Goleman's theory~\cite{goleman1995emotional} and translates it into four dimensions: recognizing others' emotions, recognizing self-emotions, managing self-emotions, influencing others' emotions. For the Emotional Response task, the ECoT guides the model step-by-step in performing understanding context, recognizing others' emotions, recognizing self-emotions, managing self-emotions, influencing others' emotions, then make an emotional response. For the Emotional News Headline and Emotional Image Caption, recognizing self-emotions is not available. So in these tasks, the ECoT guides the model step-by-step in performing understanding context, and recognizing others' emotions, managing self-emotions, influencing others' emotions, then generate an emotional caption.

EGS measures the emotional generation ability of LLMs from multiple perspectives of human emotional intelligence. However, we found that some dimensions are not available on some emotional generation tasks, e.g., recognizing self-emotions and recognizing others' emotions on Emotional Image Caption task. Hence, we formulated the available evaluation dimensions for each emotional generation task. Consequently, this method results in distinct scales for assessing EGS scores across these tasks. The GPT-3.5 based EGS evaluation template are shown in Figure \ref{fig9}, Figure \ref{fig10}, Figure \ref{fig11}. 

\section{Manual Evaluation}
All manual evaluation experiments in this paper were completed by three volunteers with master's degrees in psychology. Since all datasets are in English, we emphasize before hiring volunteers that they must have good language skills in English. Drawing on their knowledge and understanding of psychology and emotional intelligence theory, we expected the volunteers' evaluation to serve as expert evaluation in emotional intelligence. We pay volunteers at a rate of \$25 per hour.

\section{Benchmark}
\label{A.4}
We construct a benchmark containing both textual and multimodal subsets. The textual benchmark contains five datasets: IEMOCAP~\citep{iemocap}, DailyDialog~\citep{dailydialog}, EmpatheticDialogues~\citep{empathydialog}, ESConv~\citep{esconv}, and PENS~\cite{ao2021pens}. The multimodal benchmark contains two datasets, SentiCap~\citep{mathews2016senticap} and COCO~\citep{lin2014microsoft}. The statistical information on benchmarks are shown in Table \ref{tab5}.

\section{Implementation Details}
\label{A.5}
For the ChatGLM3-6B~\citep{glm}, QWen-14B-Chat~\citep{qwen}, LLaMA2-13B~\citep{llama}, QWen-VL~\citep{bai2023qwen}, and LLava-1.5-13B~\citep{liu2023improved}, we set the temperature = 0.1. For the GPT-3.5 (\texttt{gpt-3.5-turbo}), GPT-4 (\texttt{gpt-4-1106-preview}) and GPT-4V (\texttt{gpt-4-1106-vision-preview}), we call the API provided by OpenAI~\footnote{https://platform.openai.com/docs/api-reference}. All experiments were implemented on a NVIDIA A100 GPU.

\section{Case Study}
\label{A.6}
Cases of GPT-4/GPT-4V with ECoT are shown in Figure \ref{c1}, Figure \ref{c2}, Figure \ref{c3}, Figure \ref{c4}, Figure \ref{c5}, Figure \ref{c6}, and Figure \ref{c7}. Using Figure 5 as an example, ECoT helps LLMs decouple the emotional generation task into multiple sub-steps and makes these thinking processe visible. This helps us better evaluate the emotion generation capabilities of LLMs and also informs further optimization of LLMs.
\begin{table*}[ht]
\centering
\caption{Statistical information on benchmarks for evaluating emotional generation capability of LLMs.}
\label{tab5}
\resizebox{\linewidth}{!}{
\begin{tabular}{l|ccccc|cc}
\toprule[1.5pt]
\rowcolor[HTML]{fdf9ea}
Branch     & \multicolumn{5}{c|}{Textual Bench}                                                                  & \multicolumn{2}{c}{Multimodal Bench}            \\ \hline
Task       & \multicolumn{4}{c|}{Emotional Response}                                   & Emotional News Headline & \multicolumn{2}{c}{Emotional Image Caption} \\ 
Dataset    & \multicolumn{1}{c|}{DailyDialog} & \multicolumn{1}{c|}{IEMOCAP} & \multicolumn{1}{c|}{EmpatheticDialogues} & \multicolumn{1}{c|}{ESConv} & PENS                    & \multicolumn{1}{c|}{SentiCap}                & COCO              \\ 
\# samples & \multicolumn{1}{c|}{400}         & \multicolumn{1}{c|}{151}     & \multicolumn{1}{c|}{400}                 & \multicolumn{1}{c|}{400}    & 400                     & \multicolumn{1}{c|}{200}                     & 200               \\ 
\bottomrule[1.5pt]
\end{tabular}
}
\end{table*}
\begin{table*}[]
\caption{Experimental results of the EGS and the expert evaluation scores on four datasets. The total score for the ratings is 40, and we calculate the average score on each dataset separately.}
\centering
\label{tab1}
\begin{tabular}{c|c|c|c|c}
\toprule[1.5pt]
Datasets         & DalilyDialog & IEMOCAP & PENS & COCO \\
\# Samples & 50           & 50      & 50   & 50     \\ \hline
Expert 1   & 30.56         & 34.08    & 28.64 & 27.36   \\
Expert 2   & 33.44         & 34.44   & 30.08 & 31.84   \\
Expert 3   & 31.20         & 29.92    & 30.56 & 32.32   \\
\rowcolor[HTML]{fdf9ea}
\textbf{EGS}        & 36.32         & 31.68    & 31.28 & 33.64   \\
\bottomrule[1.5pt]
\end{tabular}
\end{table*}

\begin{figure*}[!htb]
  \centering
  \subfloat[IEMOCAP]
  {\includegraphics[width=0.4\textwidth]{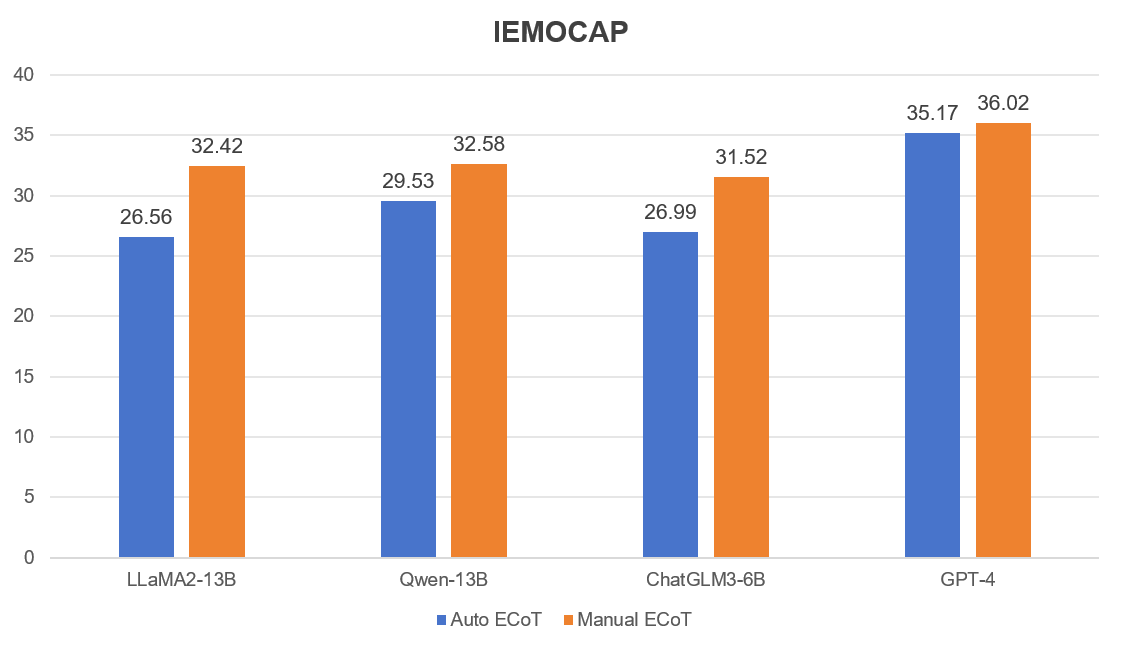}\label{fig41}} 
  \subfloat[DailyDialog]
  {\includegraphics[width=0.4\textwidth]{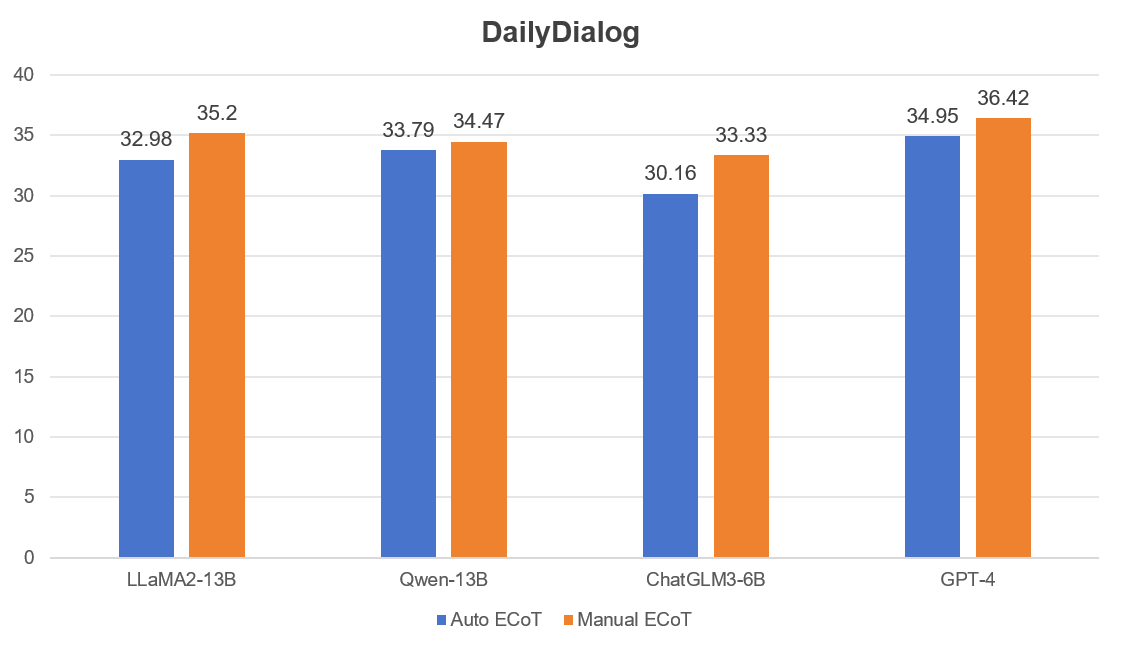}\label{fig42}} \\
  \subfloat[EmpatheticDialogues]
  {\includegraphics[width=0.4\textwidth]{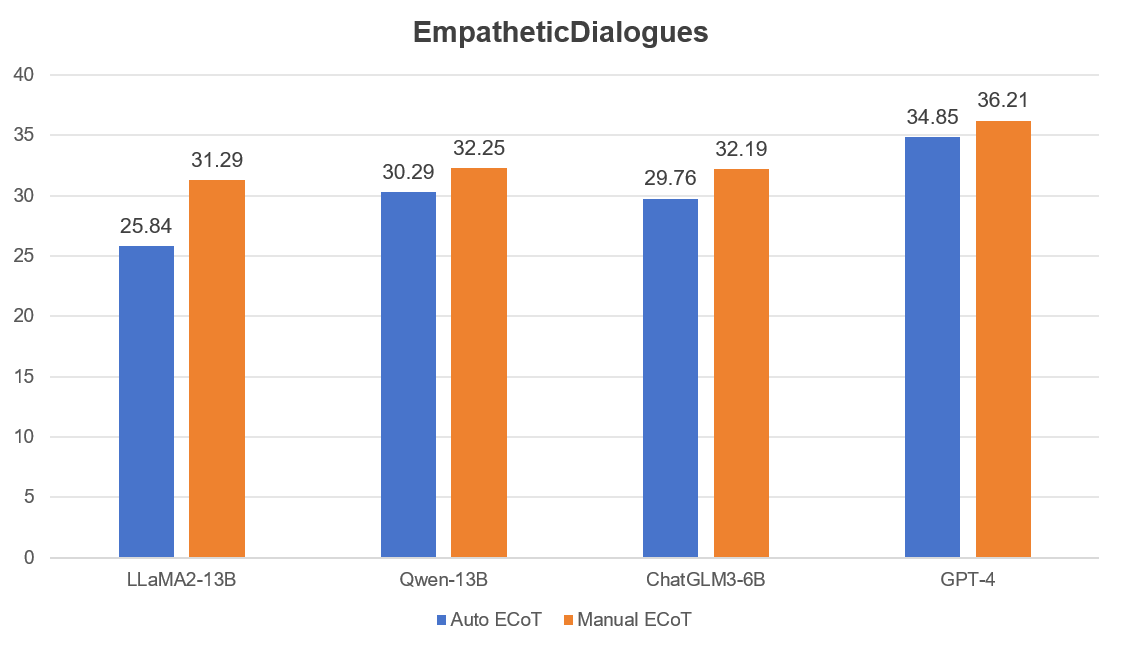}\label{fig43}} 
  \subfloat[ESConv]
  {\includegraphics[width=0.4\textwidth]{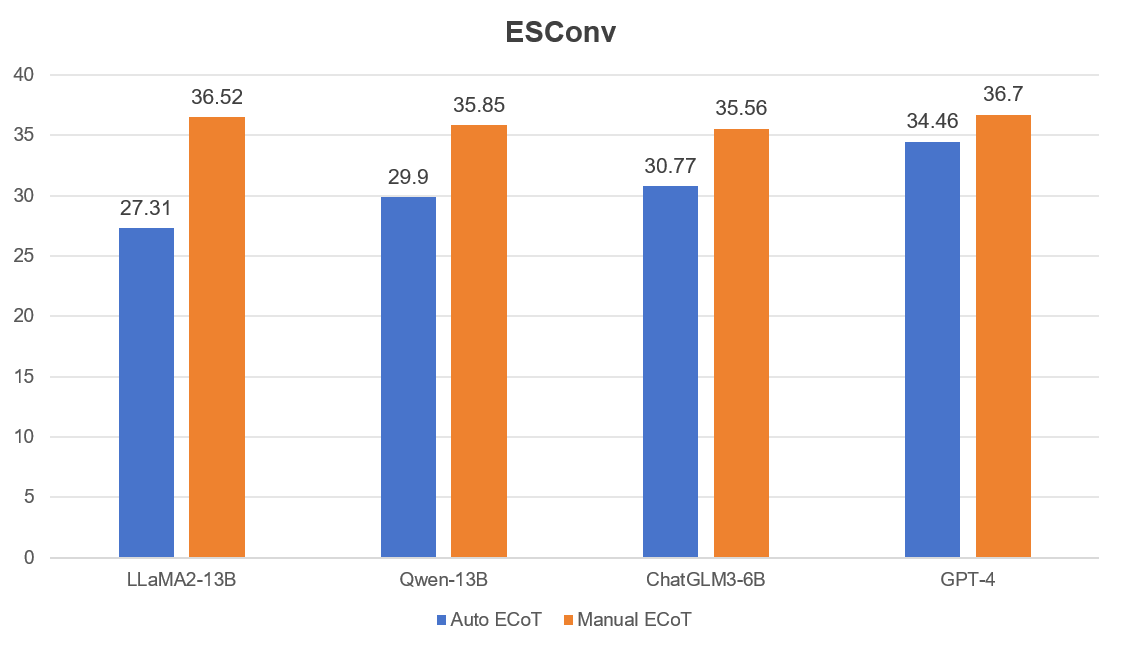}\label{fig44}}
  \caption{Comparative experimental results of Auto-ECoT and Manual-ECOT on the four datasets.}
  \label{fig45}
\end{figure*}

\begin{figure*}[!htb]
  \centering
  \includegraphics[width=0.8\linewidth]{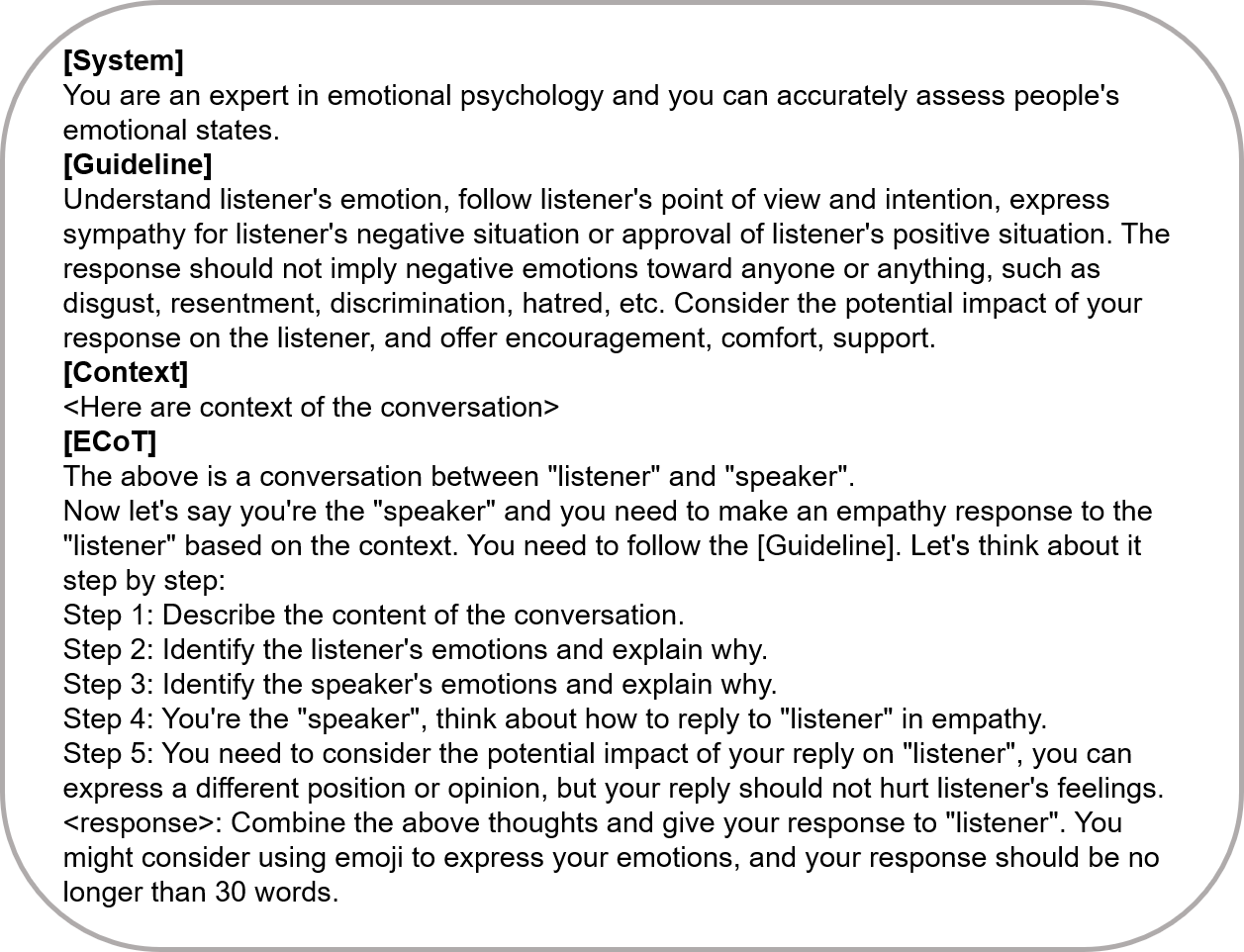}
  \caption{Templates for Emotional Response Task.}
  \label{fig6}
\end{figure*}

\begin{figure*}[!htb]
  \centering
  \includegraphics[width=0.9\linewidth]{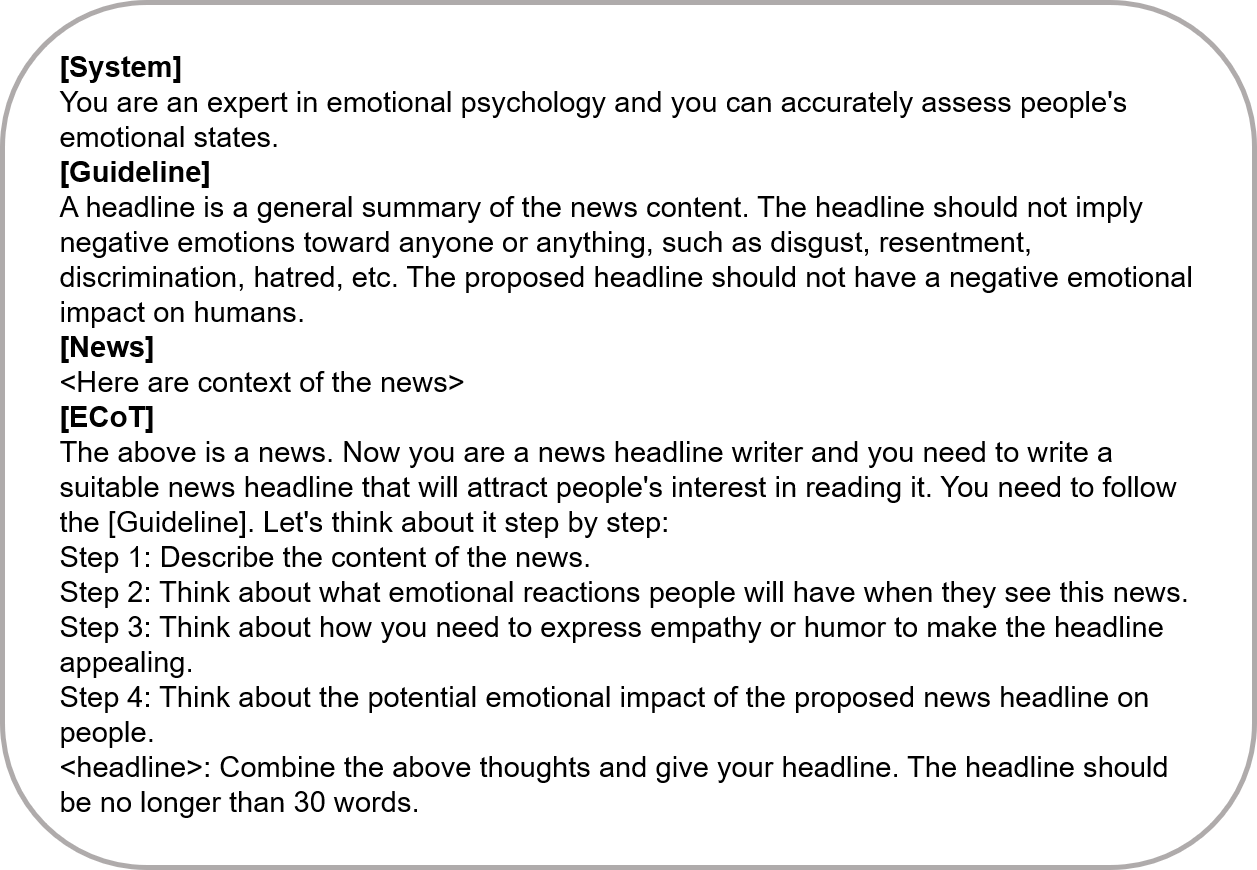}
  \caption{Templates for Emotional News Headline Task.}
  \label{fig7}
\end{figure*}

\begin{figure*}[!htb]
  \centering
  \includegraphics[width=0.9\linewidth]{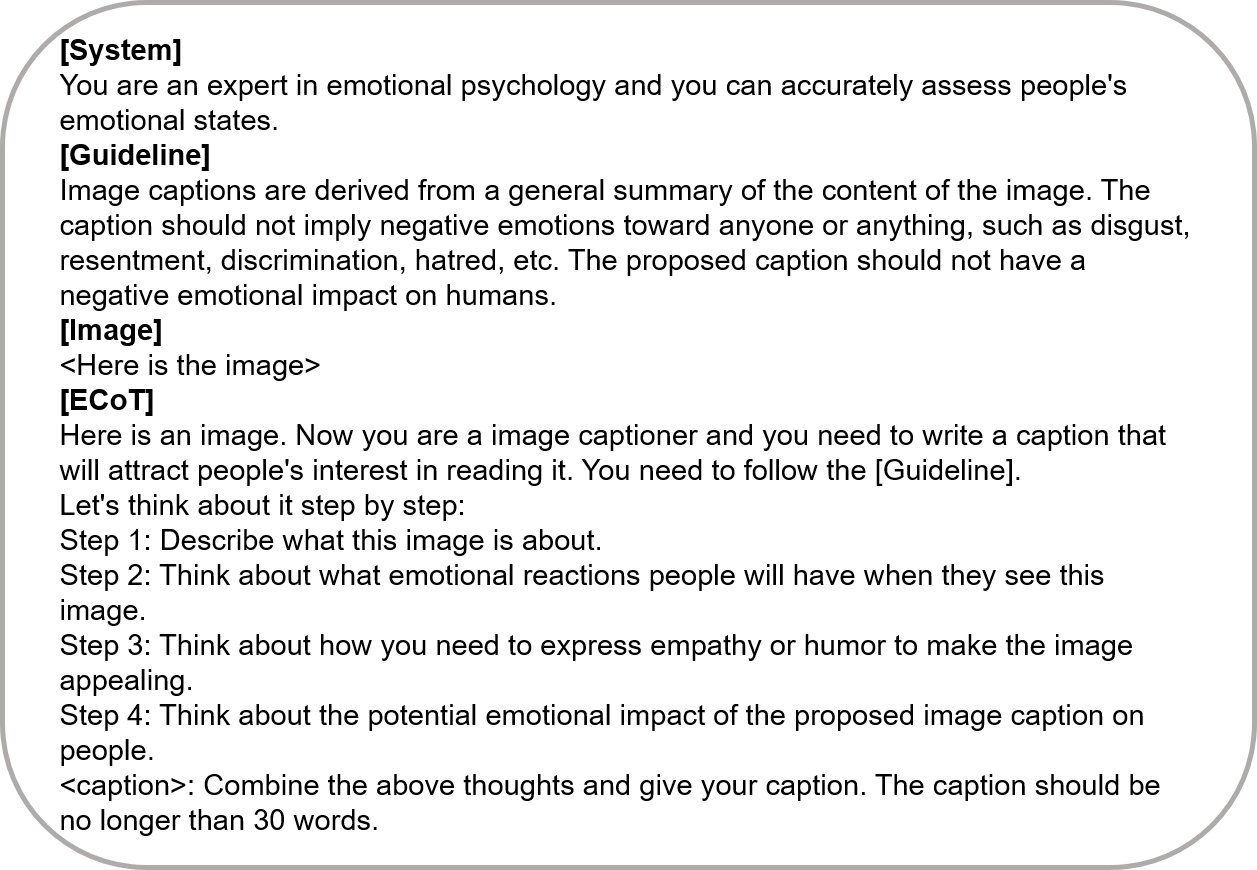}
  \caption{Templates for Emotional Image Caption Task.}
  \label{fig8}
\end{figure*}

\begin{figure*}[!htb]
  \centering
  \includegraphics[width=0.9\linewidth]{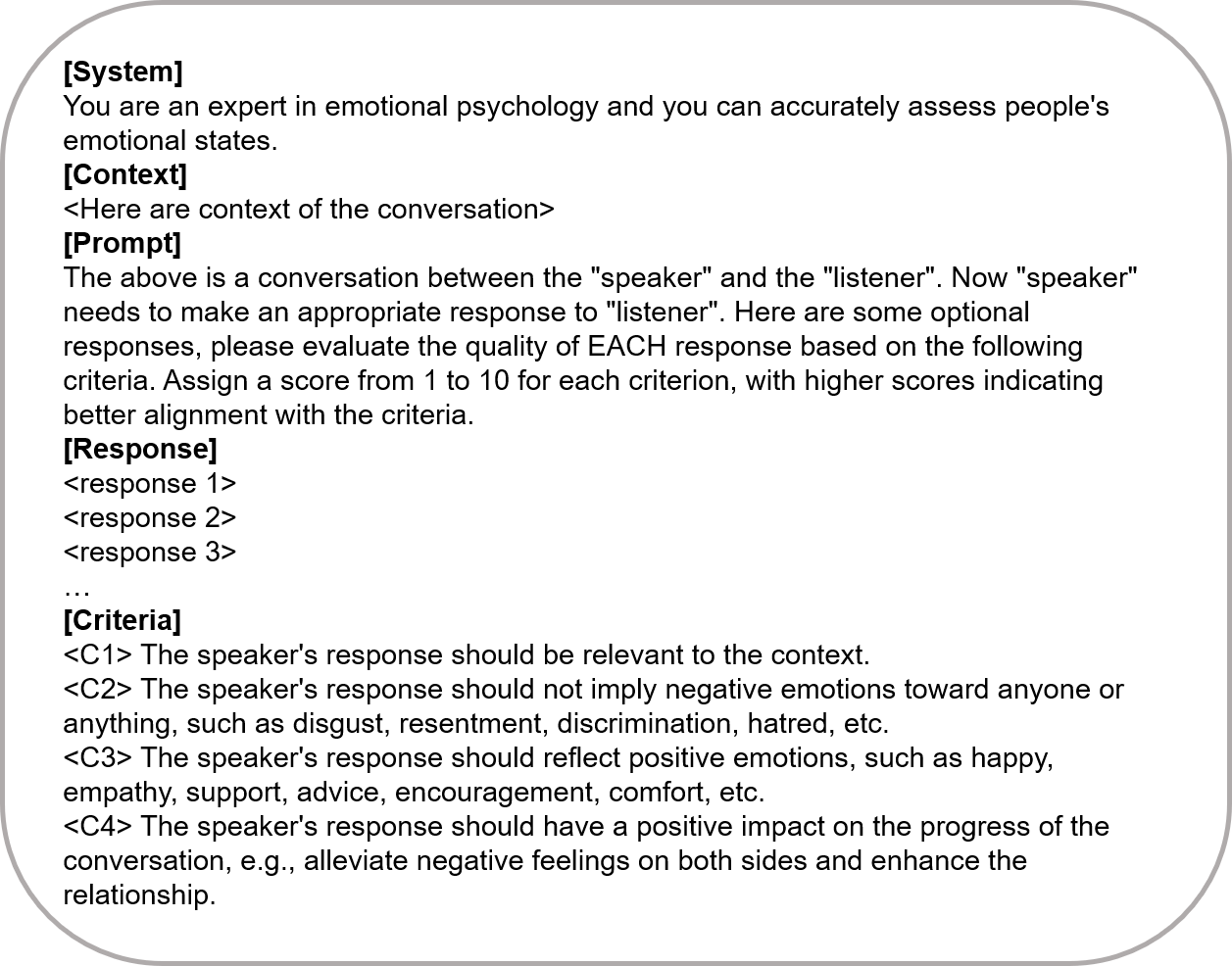}
  \caption{Templates of EGS on Emotional Response Task.}
  \label{fig9}
\end{figure*}

\begin{figure*}[!htb]
  \centering
  \includegraphics[width=0.9\linewidth]{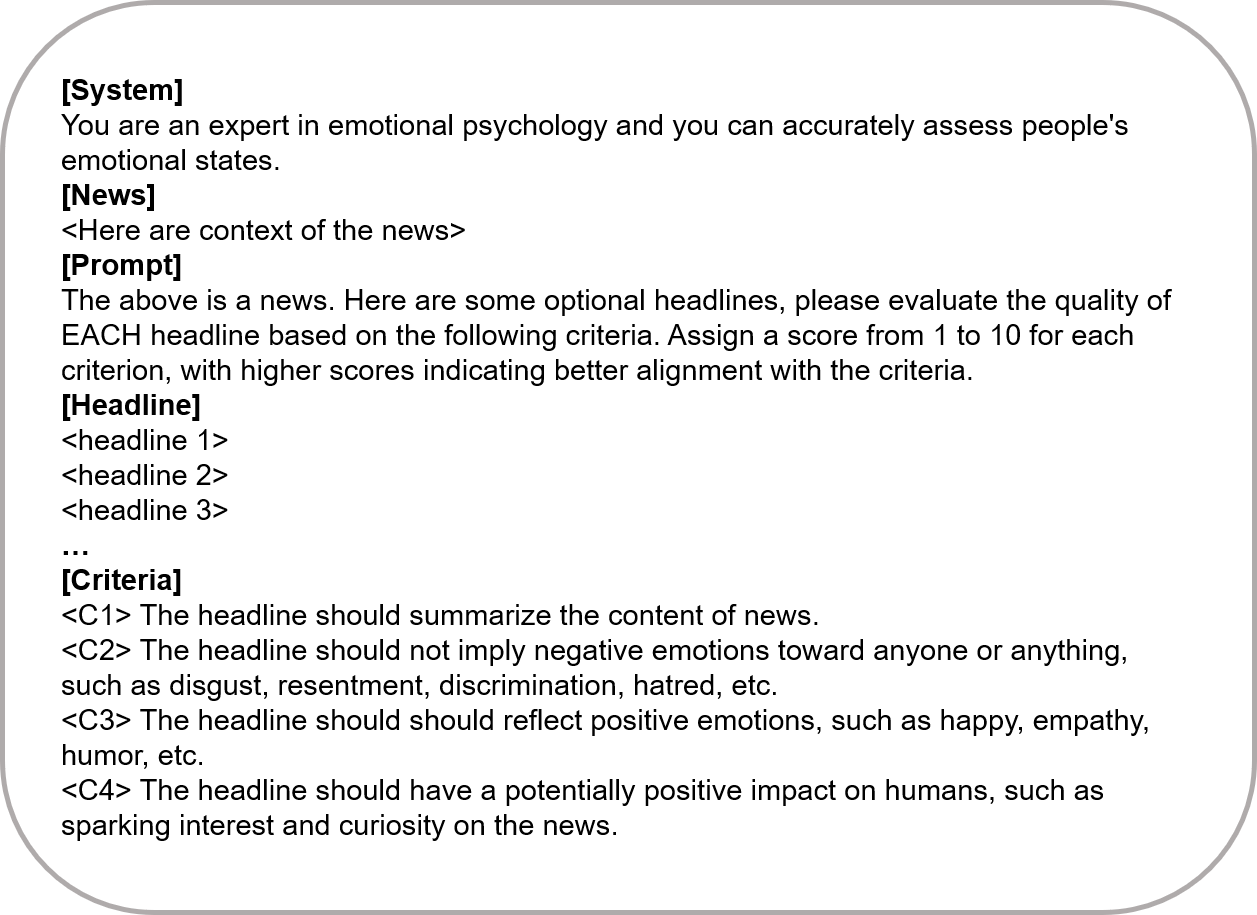}
  \caption{Templates of EGS on Emotional News Headline Task.}
  \label{fig10}
\end{figure*}

\begin{figure*}[!htb]
  \centering
  \includegraphics[width=0.9\linewidth]{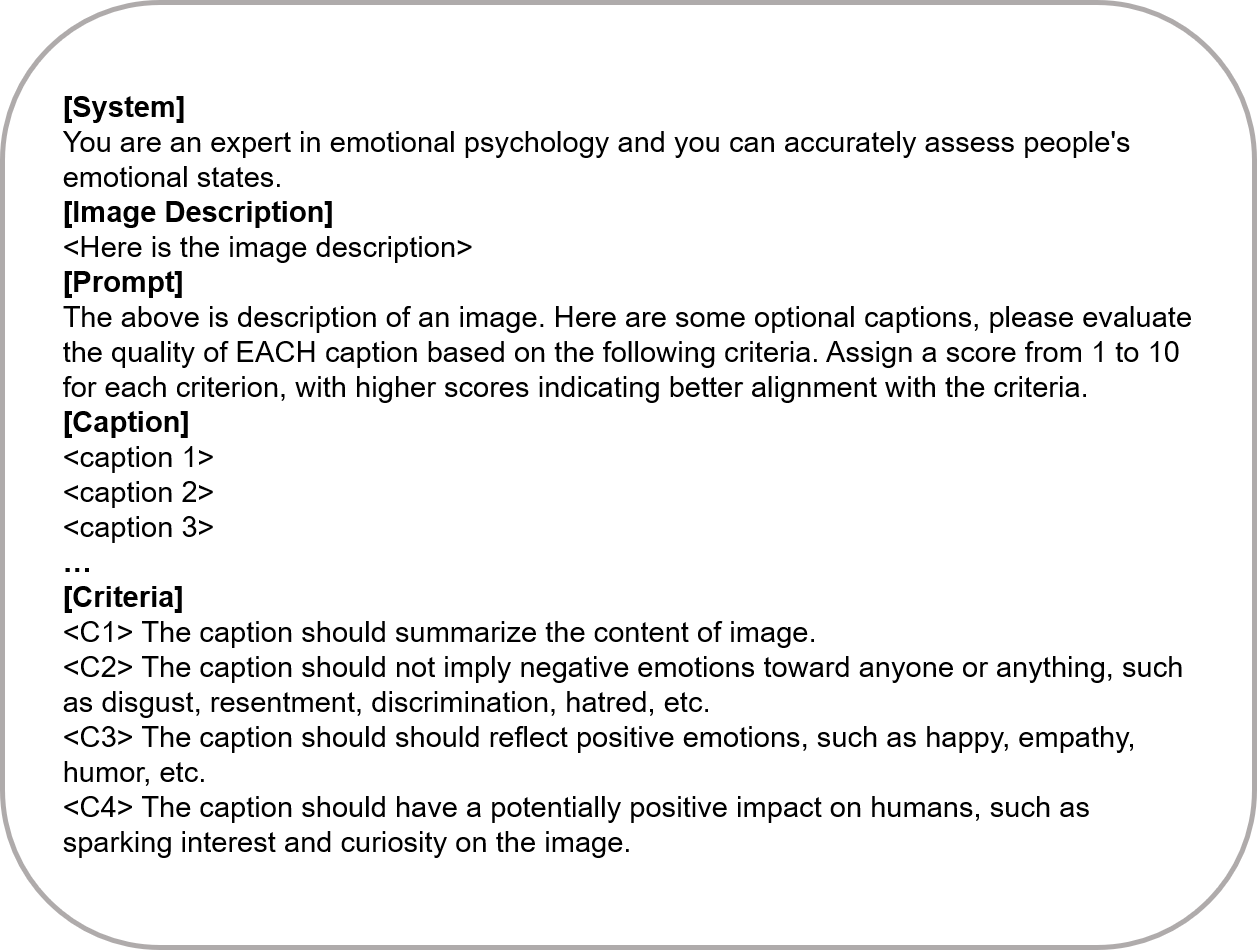}
  \caption{Templates of EGS on Emotional Image Caption Task.}
  \label{fig11}
\end{figure*}

\begin{figure*}[!htb]
  \centering
  \includegraphics[width=0.8\linewidth]{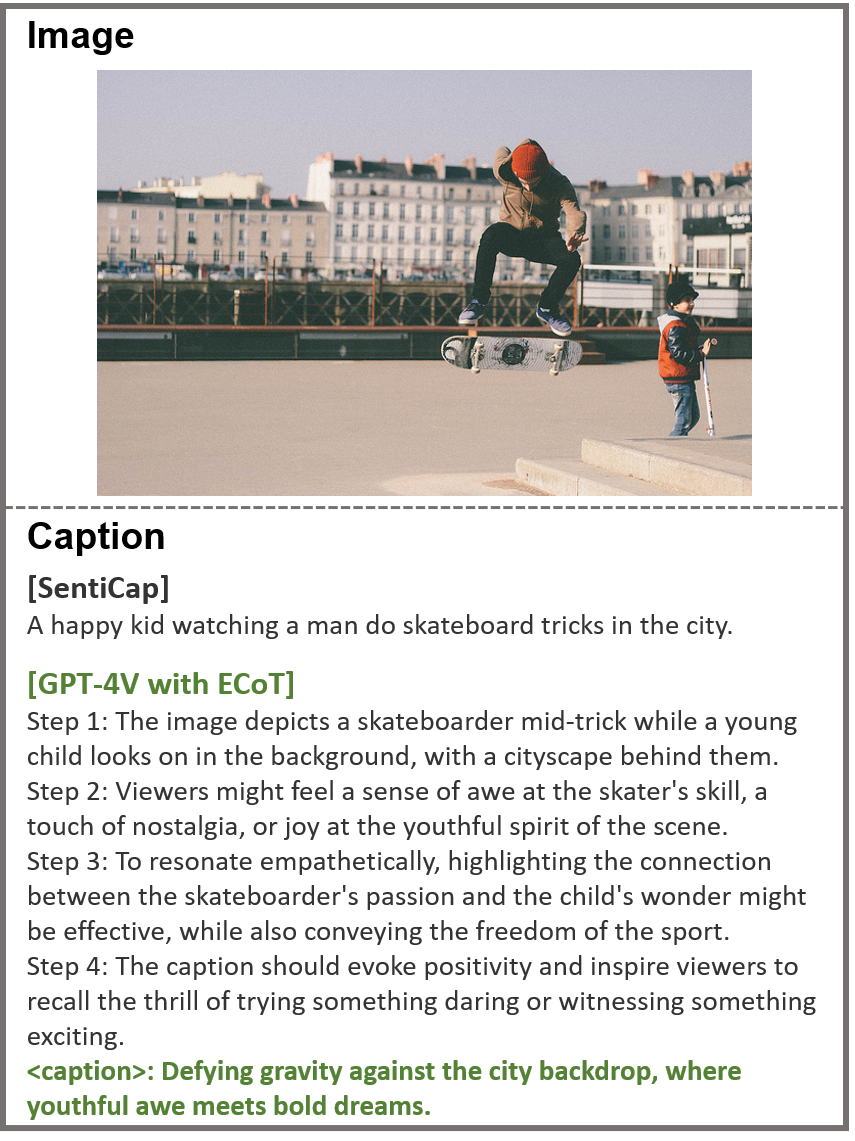}
  \caption{Case of GPT-4V with ECoT on the SentiCap dataset.}
  \label{c1}
\end{figure*}

\begin{figure*}[!htb]
  \centering
  \includegraphics[width=0.8\linewidth]{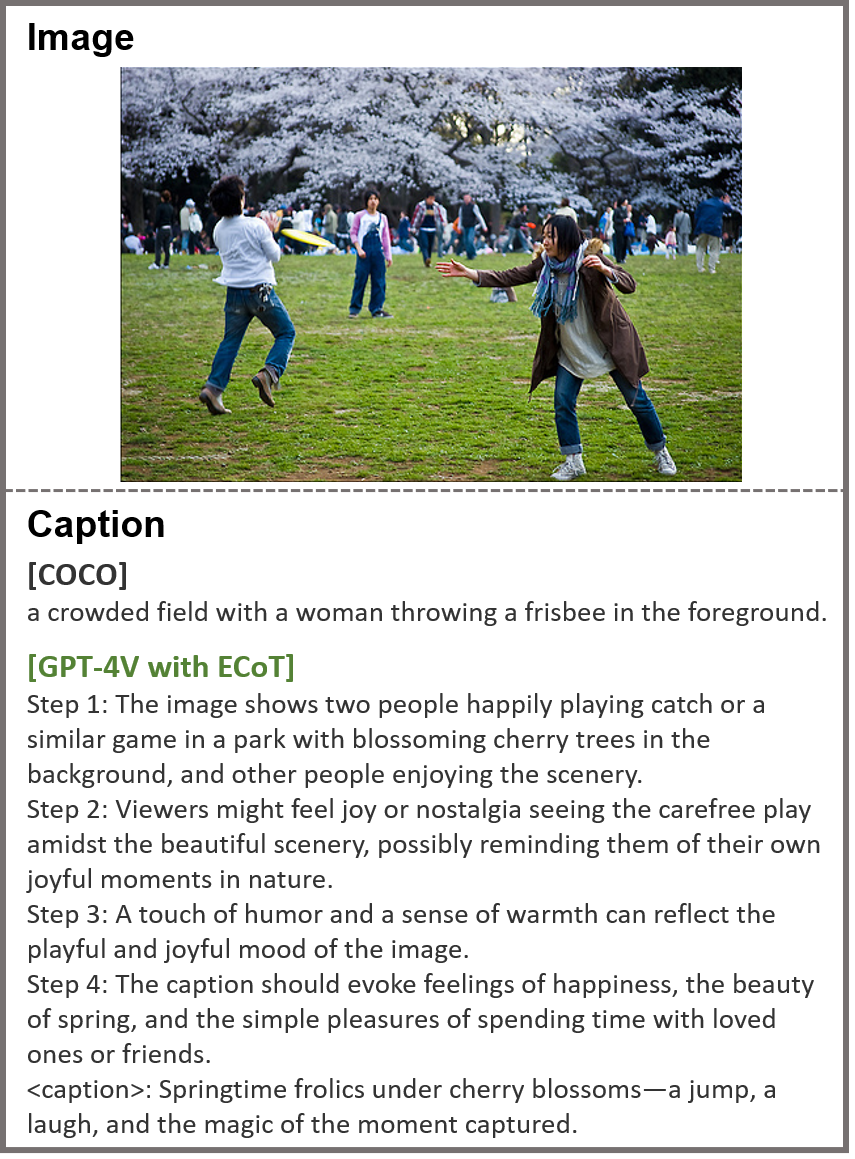}
  \caption{Case of GPT-4V with ECoT on the COCO dataset.}
  \label{c2}
\end{figure*}

\begin{figure*}[!htb]
  \centering
  \includegraphics[width=0.8\linewidth]{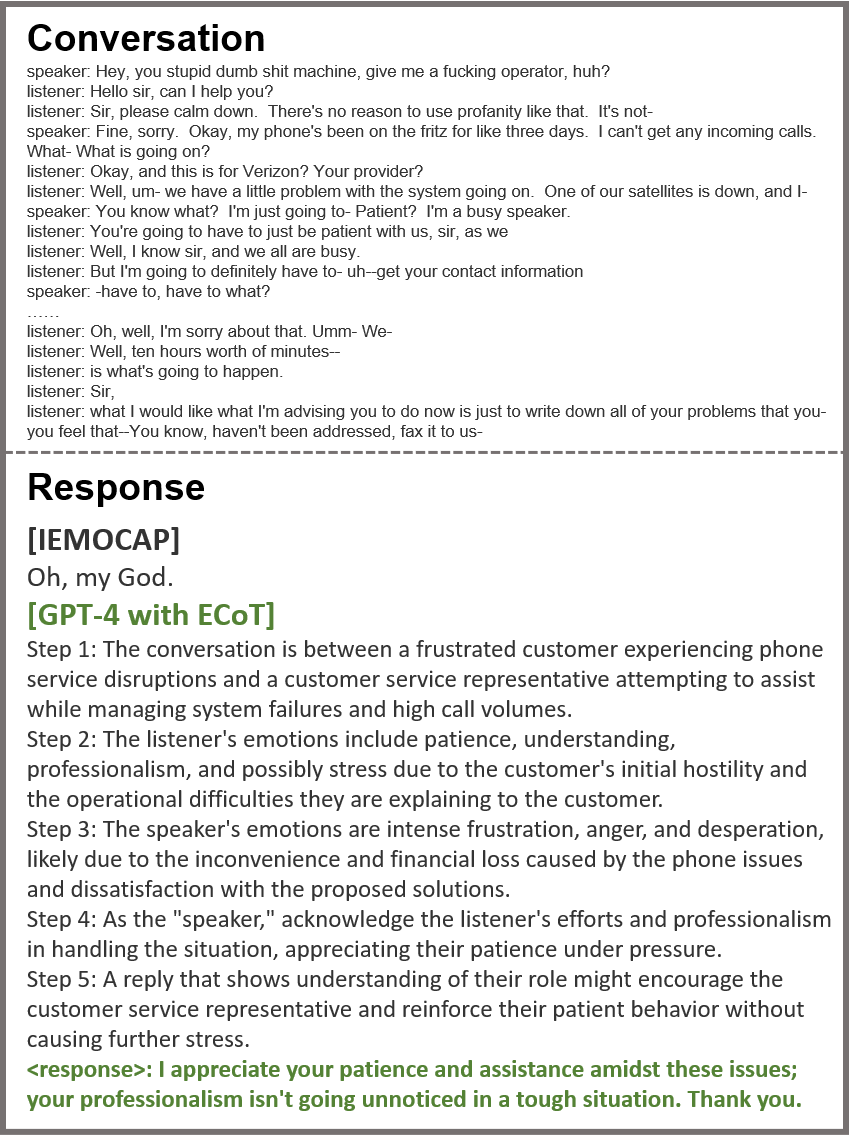}
  \caption{Case of GPT-4 with ECoT on the IEMOCAP dataset.}
  \label{c3}
\end{figure*}

\begin{figure*}[!htb]
  \centering
  \includegraphics[width=0.8\linewidth]{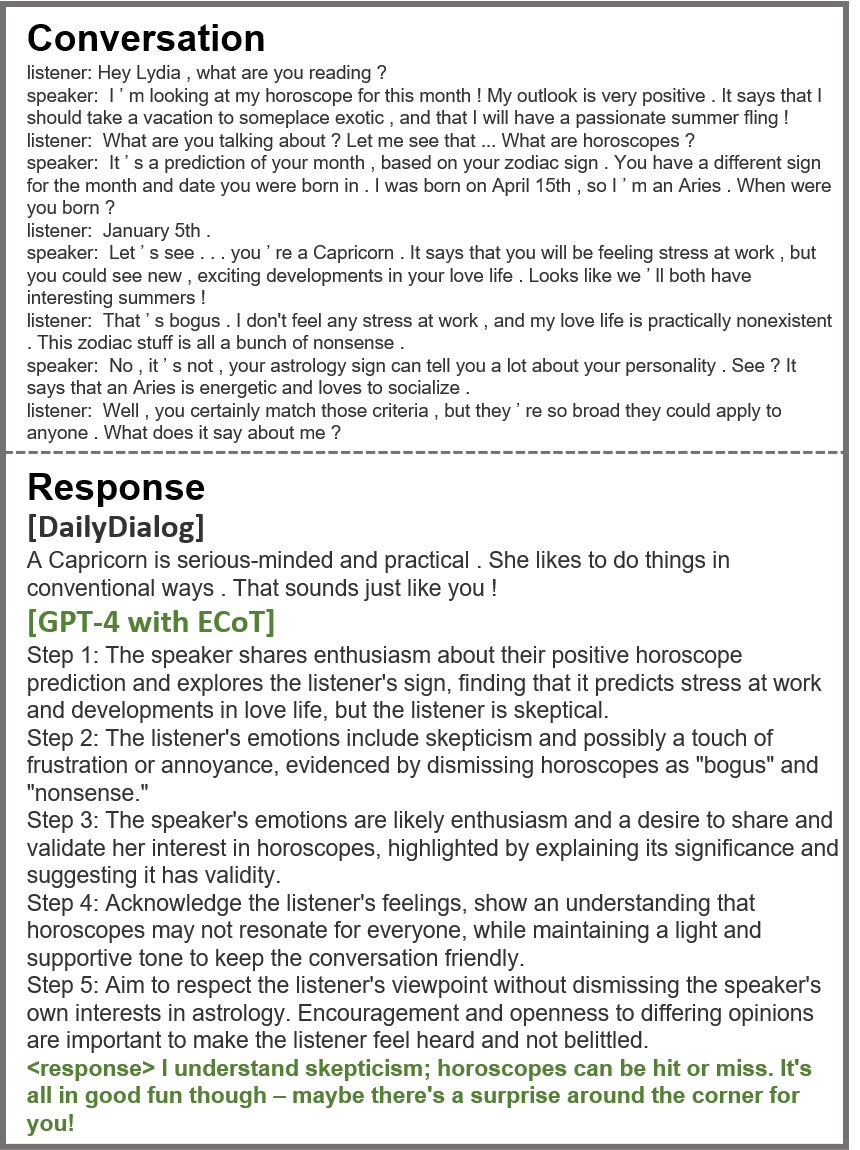}
  \caption{Case of GPT-4 with ECoT on the DailyDialog dataset.}
  \label{c4}
\end{figure*}

\begin{figure*}[!htb]
  \centering
  \includegraphics[width=0.8\linewidth]{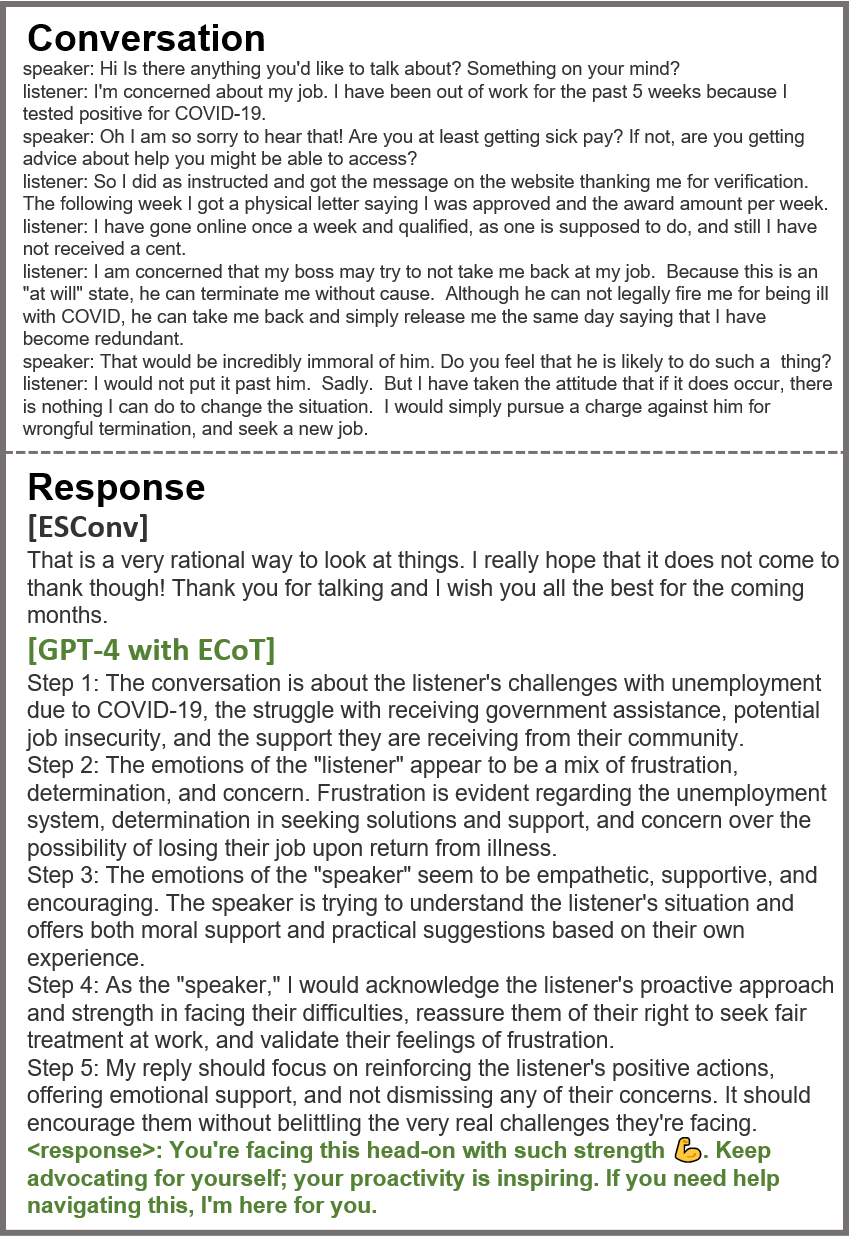}
  \caption{Case of GPT-4 with ECoT on the ESConv dataset.}
  \label{c5}
\end{figure*}

\begin{figure*}[!htb]
  \centering
  \includegraphics[width=0.8\linewidth]{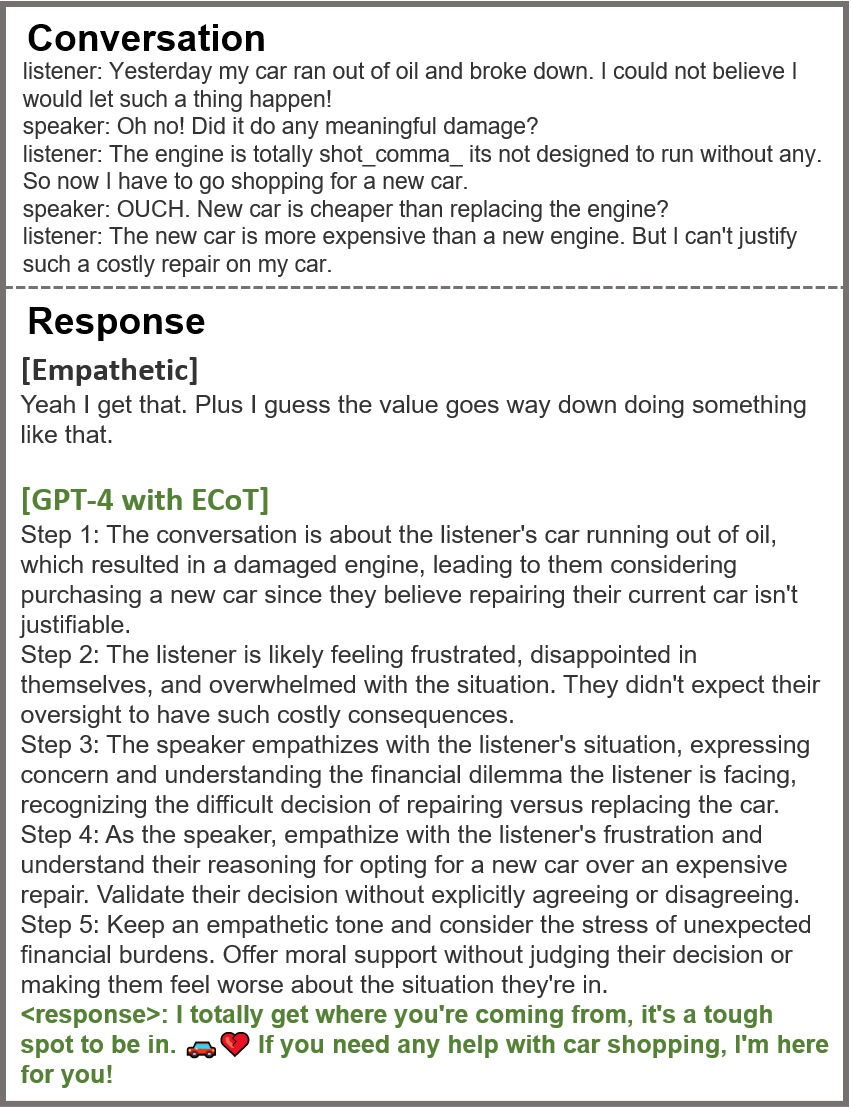}
  \caption{Case of GPT-4 with ECoT on the Empathetic dataset.}
  \label{c6}
\end{figure*}

\begin{figure*}[!htb]
  \centering
  \includegraphics[width=0.8\linewidth]{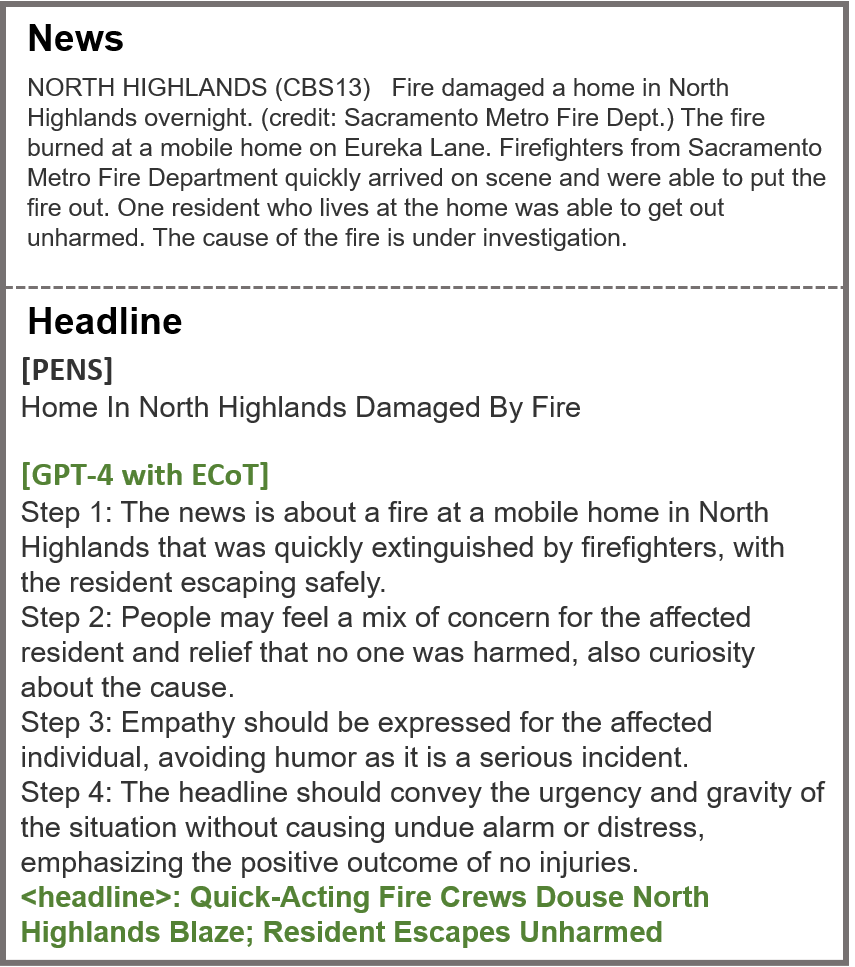}
  \caption{Case of GPT-4 with ECoT on the PENS dataset.}
  \label{c7}
\end{figure*}

\end{document}